%% file: main.tex
\documentclass[10pt,twocolumn,letterpaper]{article}
\usepackage[pagenumbers]{cvpr} 
\usepackage{graphicx}
\usepackage{amsmath}
\usepackage{algorithm}  
\usepackage{algpseudocode}  
\usepackage{float} 
\usepackage[table]{xcolor}
\usepackage{subcaption}
\usepackage{makecell}   
\usepackage{pgffor}      
\usepackage{caption}
\usepackage{tikz}
\usepackage{multirow}
\usepackage{makecell}
\usepackage{tabularray}
\usepackage[export]{adjustbox}
\usepackage{placeins}
\usepackage{afterpage}
\usepackage{colortbl}       
\usepackage{etoolbox}       
\usepackage{pgf}
\usepackage{tikz}
\usepackage[T1]{fontenc}
\usepackage{booktabs}
\usepackage{threeparttable}
\usepackage{marvosym}

\makeatletter
\def\@fnsymbol#1{\ifcase#1\or \Letter\or \dagger\or \ddagger\or
   \mathsection\or \mathparagraph\or \|\or **\or \dagger\dagger
   \or \ddagger\ddagger \else\@ctrerr\fi}
\makeatother

\definecolor{mycolor}{HTML}{9D88C8} 
\definecolor{myblue}{HTML}{B6C7EA}

\input{preamble}
\definecolor{cvprblue}{rgb}{0.21,0.49,0.74}
\usepackage[pagebackref,breaklinks,colorlinks,allcolors=cvprblue]{hyperref}

\def\paperID{10557} 
\def\confName{CVPR}
\def\confYear{2026}

\title{CORE: \textit{C}ompact \textit{O}bject-centric \textit{RE}presentations as a New Paradigm \\ for Token Merging in LVLMs}

\author{Jingyu Lei\\
Zhejiang University\\
{\tt\small jingyu.22@intl.zju.edu.cn}
\and
Gaoang Wang\thanks{Corresponding authors.}\\
Zhejiang University\\
{\tt\small gaoangwang@intl.zju.edu.cn}
\and
Der-Horng Lee\footnotemark[1]\\
Zhejiang University\\
{\tt\small dhlee@intl.zju.edu.cn}}

\begin{document}
\maketitle
\input{sec/0_abstract}    
\input{sec/1_intro}
\input{sec/2_related}

\input{sec/3_method}
\input{sec/4_experiments}

\input{sec/5_conclusion}
\clearpage

{
    \small
    \bibliographystyle{ieeenat_fullname}
    \bibliography{main}
}

\clearpage
\appendix
\setcounter{figure}{0}
\setcounter{table}{0}

\renewcommand{\thefigure}{S\arabic{figure}}
\renewcommand{\thetable}{S\arabic{table}}
\input{sec/X_suppl}

\end{document}

%% file: sec/0_abstract.tex
\begin{abstract}
Large Vision-Language Models (LVLMs) usually suffer from prohibitive computational and memory costs due to the quadratic growth of visual tokens with image resolution. Existing token compression methods, while varied, often lack a high-level semantic understanding, leading to suboptimal merges, information redundancy, or context loss. To address these limitations, we introduce CORE (\textbf{C}ompact \textbf{O}bject-centric \textbf{RE}presentations), a new paradigm for visual token compression. CORE leverages an efficient segmentation decoder to generate object masks, which serve as a high-level semantic prior to guide the merging of visual tokens into a compact set of object-centric representations. Furthermore, a novel centroid-guided sorting mechanism restores a coherent spatial order to the merged tokens, preserving vital positional information. Extensive experiments show that CORE not only establishes a new state-of-the-art on six authoritative benchmarks for fixed-rate compression, but also achieves dramatic efficiency gains in adaptive-rate settings. Even under extreme compression, after aggressively retaining with only 2.2\% of all visual tokens, CORE still maintains 97.4\% of baseline performance. Our work demonstrates the superiority of object-centric representations for efficient and effective LVLM processing.

\end{abstract}

%% file: sec/1_intro.tex
\section{Introduction}

By deeply integrating visual perception with language intelligence, Large Vision-Language Models (LVLMs) \citep{bai2023qwenvlversatilevisionlanguagemodel,chen2023sharegpt4vimprovinglargemultimodal,li2023blip,li2024minigeminiminingpotentialmultimodality,liu2024improvedbaselinesvisualinstruction} represent a milestone step towards Artificial General Intelligence (AGI). However, visual tokens produced by a Vision Transformer (ViT) increase quadratically with the input image resolution \citep{shao2025tokenstalkmuchsurvey}. For instance, an image of size 1024×1024 pixels generates 4096 tokens with a 16×16 patch size. This large token count imposes prohibitive computational and memory costs on the downstream Large Language Models (LLMs) \citep{openai2024gpt4technicalreport,bai2023qwentechnicalreport,touvron2023llamaopenefficientfoundation,zhu2023minigpt4enhancingvisionlanguageunderstanding,brown2020language,radford2019language,touvron2023llama2openfoundation}, as the complexity of self-attention is quadratic (\(O(N^2)\)) with respect to the sequence length \citep{liang2022not,vaswani2017attention}. 

\begin{figure}[t]
    \centering 
    \begin{subfigure}[t]{0.48\linewidth}
        \centering
        \includegraphics[width=\linewidth]{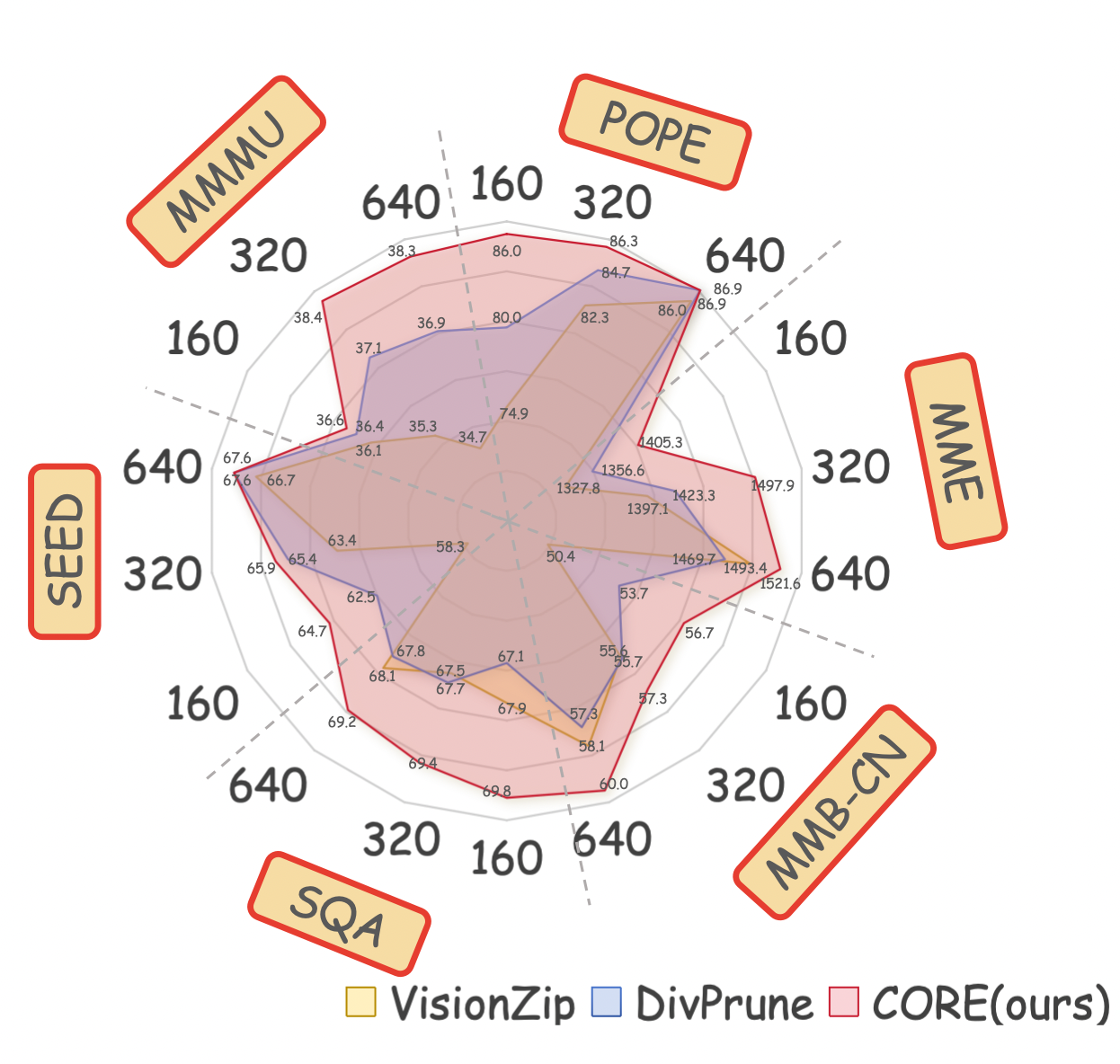} 
        \caption{Performance on 6 benchmarks.}
        \label{CORE achieves sota results.} 
    \end{subfigure}
    \begin{subfigure}[t]{0.48\linewidth}
        \centering
        \includegraphics[width=\linewidth]{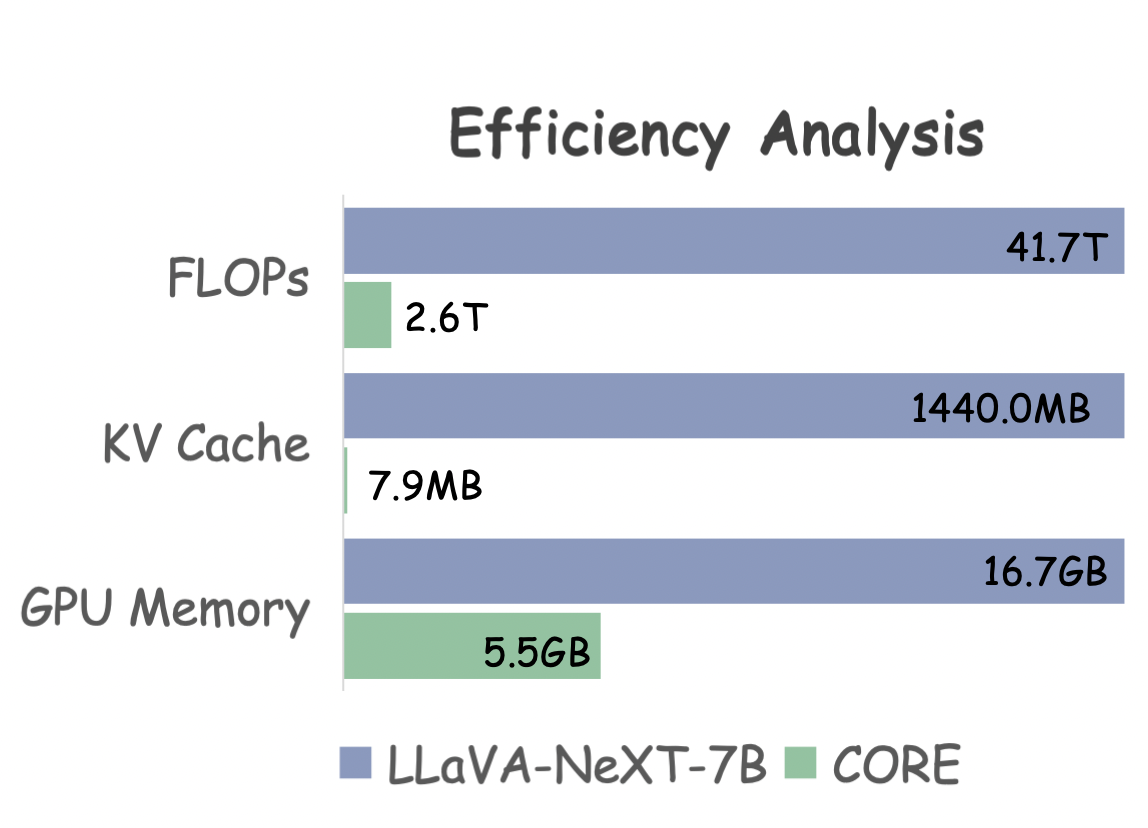} 
        \caption{Efficiency Analysis.}
        \label{Efficiency Analysis} 
    \end{subfigure}

    \caption{\textbf{CORE's Performance and Efficiency.} (a) When retaining with only 160, 320, and 640 tokens, CORE outperforms current state-of-the-art efficient LVLMs, such as VisionZip \citep{yang2024visionzip} and DivPrune \citep{alvar2025divprune}, across six benchmarks. (b) Under its highest compression ratio, CORE reduces FLOPs by 16.0\(\times\), KV Cache by 182.3\(\times\), and GPU Memory by 2.7\(\times\), while still maintaining 97.4\% of its baseline performance. }
    \label{CORE's Performance and Efficiency} 
\end{figure}

To address this challenge, a variety of visual token compression methods \citep{ma2025shortlvlmcompressingacceleratinglarge,yang2025visionthinksmartefficientvision,liu2025tokenpruningoperationpruning,kong2025clappercompactlearningvideo,qi2025lmmefficientvideounderstanding,Dhouib_2025_CVPR,lu2025internvlxadvancingacceleratinginternvl,zhang2025vqtokenneuraldiscretetoken,li2025fcotvladvancingtextorientedlargevisionlanguage,yang2025pvc} have been proposed. Existing token compression methods are generally based on transformation, similarity \citep{shen2025fastviddynamicdensitypruning,li2024videochat,han2025filtercorrelatecompresstrainingfree,hyun2025multi}, attention \citep{huang2025dynamicllavaefficientmultimodallarge,li2025redundancylensrevealingexploitingvisual,liang2025dynamictokenreductiongeneration,ye2025fit} and query \citep{luo2025vcm,liu2025hybrid,Ye_2025_CVPR}. These methods start from individual tokens and can be categorized as \textit{token-centric} methods. Some token-centric methods merge tokens with similar features \citep{zhang2025dyntokdynamiccompressionvisual,xu2025auroralongbringingrnnsefficient}, but these tokens are not always redundant. 
Consequently, tokens with merely similar texture might be merged together, potentially confusing two semantically distinct entities. Other methods concentrate on tokens with high attention scores \citep{sun2025adatpattentiondebiasedtokenpruning,guo2025filavideospatiotemporalcompressionfinegrained,shao2025twigvlm}, which cannot guarantee that the retained tokens are compact and not redundant among themselves. Still other methods \citep{zhang2025beyond} bind the compression process to specific text queries and retain the more relevant tokens. This approach, however, comes at the cost of the model's generality, undermining its comprehension of the complete scene context. Despite recent progress, existing token-centric token compression strategies often face a common challenge: operating without a high-level, semantic understanding of the scene.

Motivated by the limitations of prior works, we introduce CORE (\textbf{C}ompact \textbf{O}bject-centric \textbf{RE}presentations), a new paradigm wherein each distinct object is consolidated into a single and compact token for the LLM. 
CORE consists of a ConvNeXt-L \citep{liu2022convnet} backbone, a Mask2Former \citep{cheng2021mask2former} segmentation head and an LLM language head.
To be specific, CORE routes features from the ConvNeXt-L visual encoder to the internal Mask2Former segmentation head to generate object masks. These masks then guide an object-centric merging of the visual tokens. The resulting compact tokens are subsequently spatially sorted and fed into the LLM decoder. This process is highly efficient, as a shared visual encoder provides features for both the segmentation and language decoding pathways, significantly minimizing computational overhead.
CORE's paradigm largely resolves the aforementioned issues by generating an ordered set of compact object-centric representations for the entire scene. By leveraging semantic priors, it effectively prevents the merging of semantically distinct but texturally similar tokens. This compact representation also eliminates the intra-object redundancy common in attention-based methods. Furthermore, unlike query-bound approaches, CORE preserves the complete scene context, providing a more robust foundation for complex tasks.
As shown in Fig.~\ref{CORE achieves sota results.}, CORE achieves state-of-the-art performance across six authoritative image understanding benchmarks at three compression rates. Fig.~\ref{Efficiency Analysis} shows CORE brings substantial efficiency improvements on the dynamic adaptive compression tasks. Our main contributions are summarized as follows:

\begin{itemize}
\item We introduce CORE, a new paradigm for LVLMs that pioneers object-centric token merging, creating compact representations that provide enhanced semantic clarity as well as breakthrough efficiency.

\item CORE leverages an end-to-end architecture built upon a shared visual encoder, which provides features for both the segmentation and the language head, largely mitigating the additional segmentation computational overhead.

\item CORE not only outperforms state-of-the-art efficient LVLMs on fixed-rate compression tasks, but also drastically reduces computational and memory costs in adaptive-rate compression scenarios with negligible performance degradation.

\end{itemize}

%% file: sec/2_related.tex
\section{Related Work}

\subsection{Transformation-based Token Compression} 

The most straightforward methods \citep{liu2025lacoefficientlayerwisecompression,tang2025learningcompactvisiontokens,jiang2025stormtokenefficientlongvideo,fastvlm2025} compress tokens via mathematical or algorithmic transformations, such as pixel unshuffle, spatial pooling, interpolation and convolution, which preserve the spatial locality of 2D features. While these methods \citep{li2024llavaonevisioneasyvisualtask,cai2024matryoshkamultimodalmodels,yao2024decodecouplingtokencompression,cha2023honeybee} are simple and efficient, their compression mechanism is inherently blind and inflexible, incapable of dynamically identifying and preserving key features.

\subsection{Similarity-based Token Compression} 
Similarity-based methods \citep{sun2025llava,jeddi2025similarity,chai2024auroracap,wang2025folderacceleratingmultimodallarge} posit that similar tokens contain similar information and cause redundancy. Specifically, this is achieved by calculating the pairwise distances or similarities between visual tokens and merging similar tokens \citep{wang2025dymudynamicmergingvirtual,jeddi2025similarityawaretokenpruningvlm} or retaining tokens with the greatest difference \citep{ma2025mmgvidmaximizingmarginalgains,cho2025floc}. ToMe \citep{bolya2022token} is a typical method, which proposes the bipartite soft matching algorithm to identify and merge visual tokens inside vision transformers (ViTs). 
TopV \citep{yang2025topv} builds an optimization problem that considers a combination of factors, including feature similarity and spatial location, in order to select a visual token subset that is both representative and concise. A problem with such methods is that decisions based purely on local feature affinity are semantically blind and prone to erroneous merges between semantically distinct but texturally similar regions.

\subsection{Attention-based Token Compression} 
Attention-based token compression methods \citep{sun2025adatp,liu2025compression,zhuang2025st3,zhang2025vispruner,hu2025lightvlmaccelerainglargemultimodal,zhu2025visionselectorendtoendlearnablevisual} use attention scores as a direct proxy for token importance. They operate on the assumption that tokens with low attention scores are redundant and can be pruned with minimal impact on model performance. Such compression can occur either in the vision encoder \citep{liu2025compression,arif2025hired} or the LLM decoder \citep{ye2024atpllavaadaptivetokenpruning,he2024zipvlefficientlargevisionlanguage,zhang2024pmod}. HiPrune \citep{liu2025hiprune} selects an information-rich and diverse subset of visual tokens by hierarchically analyzing the attention scores within the vision encoder. VTW \citep{lin2025boosting} observes that visual tokens in deeper layers garner very little attention and withdraws them beyond a certain predetermined layer to speed up inference. 
While this strategy ensures that the retained tokens are salient, it does not guarantee they are non-redundant, thus failing to address the problem of information redundancy. A more critical challenge arises with decoder-side pruning: it requires access to attention scores that are never explicitly computed by modern acceleration libraries like FlashAttention \citep{dao2022flashattention,dao2023flashattention}. 

\subsection{Query-based Token Compression}

\begin{figure*}[t]
\centering
\includegraphics[width=0.99\textwidth]{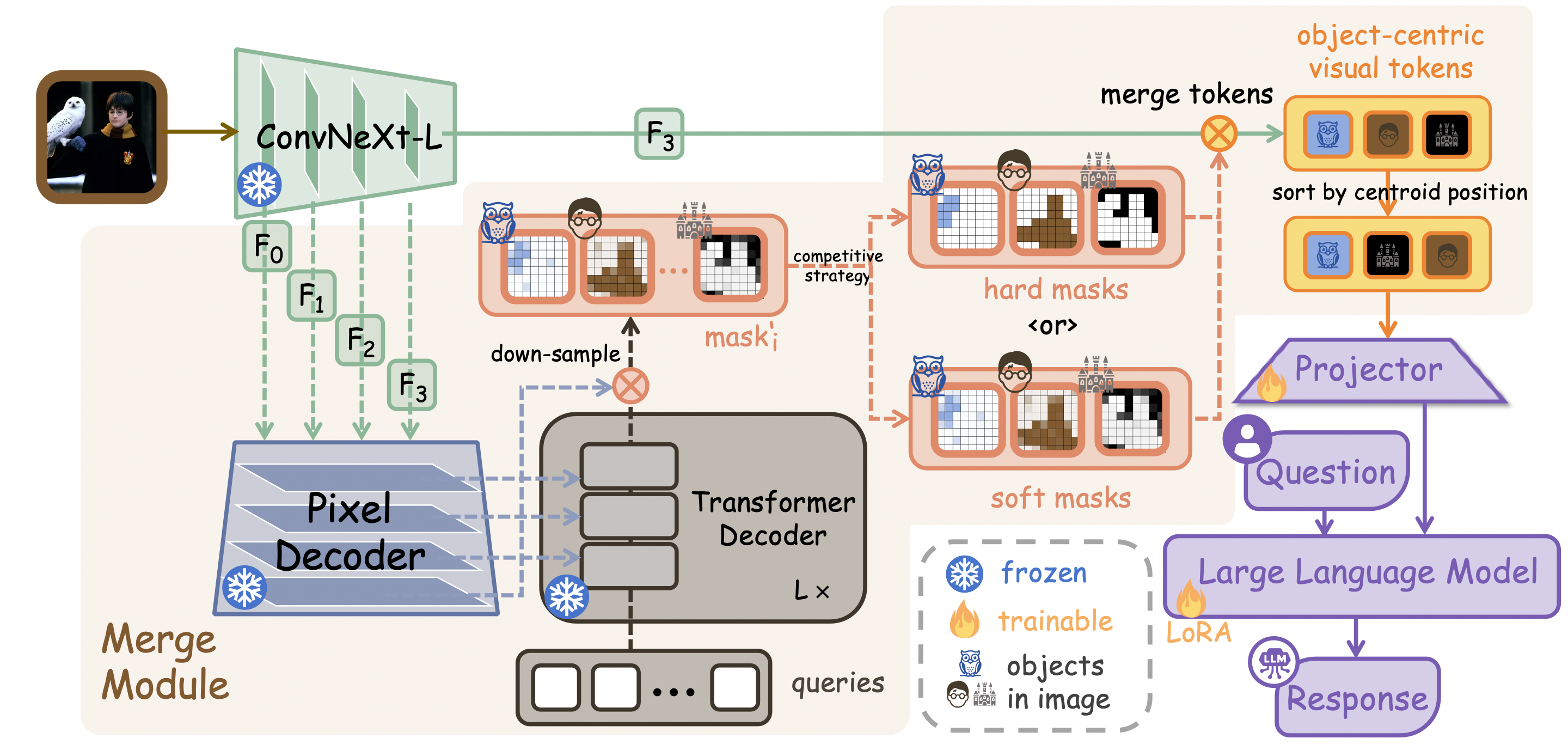}
\caption{\textbf{Overview of CORE.} Our framework consists of two key pathways. The primary data flow, indicated by solid lines, shows how compact object-centric representations are generated and processed by the language decoder. This process is informed by the auxiliary segmentation head, shown with dashed lines, which produces the object masks that guide the token merging. The icon in the top-left corner of each mask denotes a different object in the image.}
\label{Overview of CORE}
\end{figure*}

To enhance computational efficiency, query-based compression methods \citep{zeng2025glimpse,zhang2025top,shi2025staticdynamicqueryadaptivetoken} leverage textual relevance to guide the selective reduction of visual tokens, thereby preserving only the information most relevant to a specific task. This principle is realized in two main ways. Explicit methods use the user's query as a filter to extract the most pertinent visual tokens \citep{li2025qgvtcquestionguidedvisualtoken,han2025adafvrethinkingvisuallanguagealignment}, while implicit methods distill the visual information into a fixed number of text-relevant tokens \citep{yan2025crosslmmdecouplinglongvideo,zhang2025falcon,llavamini,li2023llamavidimageworth2,wen2024efficientvisionlanguagemodelssummarizing}. MMTok \citep{dong2025mmtok} employs a greedy algorithm to select a subset of visual tokens that simultaneously maximizes coverage of both the text query's semantics and the entire image's visual information. CDPruner \citep{zhang2025beyond} builds upon the deduplication concept from similarity-based methods, but makes the pruning process dynamic and intelligent by incorporating user queries as a decision criterion. 
However, these explicit approaches risk losing crucial contextual information, a problem exacerbated by ambiguous queries or in conversational contexts. The implicit approach, on the other hand, introduces a fixed information bottleneck, which may be insufficient for representing visually complex scenes.

Some methods \citep{shao2025holitomholistictokenmerging,fu2025framefusioncombiningsimilarityimportance,xiao_retake_2024,liu2024multi,zhu2024focusllavacoarsetofineapproachefficient,TokenPacker,hu2024illavaimageworthfewer} further integrate multiple of the above elements. VISA \citep{jiang2025visa} operates by first selecting key visual tokens guided by a text query, subsequently aggregating information from the non-selected tokens into the kept set based on visual similarity. GreedyPrune \citep{pei2025greedyprune} employs a joint optimization approach to ensure that the selected token subset is both semantically salient and visually diverse.

%% file: sec/3_method.tex
\section{Method}


\textbf{Motivation and Overview.} To address the prevalent issue of token-centric compression methods for LVLMs, we emulate human perception by adopting an object-centric approach to obtain a more efficient and robust compression paradigm. 
CORE, as an end-to-end architecture, leverages an \textit{intrinsic} segmentation prior to decompose the image and merge each individual object, as well as subsequently sorts these merged tokens by a centroid-guided strategy.
Fig.~\ref{Overview of CORE} shows that we achieve this goal through efficient reuse of a ConvNeXt-L \citep{liu2022convnet} vision encoder and a Mask2Former \citep{cheng2021mask2former} decoder, which includes a Pixel Decoder and a Transformer Decoder. An image is input to the ConvNeXt-L vision encoder to extract a multi-scale feature pyramid which is utilized by the Mask2Former decoder. These masks are then filtered via a competitive strategy, yielding a set of soft or hard masks, each corresponding to a distinct visual entity. Guided by these masks, the final-layer features from the \textit{same} ConvNeXt-L encoder are merged on an object-by-object basis. This set of object-centric tokens, spatially sorted by their centroids, forms the visual tokens fed to the language decoder. This shared-encoder design significantly reduces the computational overhead compared to multi-backbone approaches.

\subsection{Model Architecture}
\label{Model Architecture}

\textbf{ConvNeXt-L Backbone as Shared Vision Encoder.} The overall architecture of CORE follows LLaVA-NeXT \citep{liu2024improved,liu2023llava}. LLaVA-NeXT employs a CLIP ViT-L/14 \citep{radford2021learning}, but its single-scale feature map creates an architectural conflict with our segmentation head, which demands a multi-scale pyramid. We resolve this by replacing the ViT with a ConvNeXt-L \citep{liu2022convnet} backbone from OpenCLIP \citep{ilharco2021openclip}. As a hierarchical CNN, ConvNeXt-L innately provides the required feature pyramid. 
Besides, we utilize a ConvNeXt-L variant pretrained under the CLIP contrastive objective, ensuring its output features are aligned with the text embedding space and thus fully compatible with the LLaVA-NeXT framework. This strategic selection allows for a single, unified backbone to efficiently serve two heterogeneous downstream tasks. The encoder outputs $F_C=\{F_0,F_1,F_2,F_3\}$ at $1/4$ to $1/32$ resolutions. We route the full pyramid $\{F_0, \dots, F_3\}$ to the segmentation decoder, while the semantically-rich $F_3$ map is used as the visual input for the language decoder, as shown in Fig.~\ref{Overview of CORE}.

\textbf{Mask2Former as Segmentation Head Decoder.} We utilize Mask2Former \citep{cheng2021mask2former} to generate object masks. Mask2Former's decoders includes a Pixel Decoder and a Transformer Decoder. The Transformer decoder receives the set of multi-scale features from the Pixel Decoder and \(N\) initialized object queries. Through \(L\) layers of the Transformer Decoder, each learnable query will gradually lock onto a specific object and finally output a probability mask after applying a sigmoid function. The probability masks are then down-sampled via an interpolation function to match the resolution of \(F_3\). Subsequently, we employ a pixel-wise \textit{competitive strategy} to filter the set of \(N\) predicted masks. This approach differs from conventional confidence-based thresholding and non-maximum suppression. Specifically, we identify the mask that exhibits the highest probability score for each pixel location. Only the queries corresponding to masks that are maximal for at least one pixel are retained as valid. From this process, two distinct outputs \(\mathcal{P}_{\text{valid}}=\left\{P_1, \ldots, P_N\right\}\), where each mask \(P_n\) corresponds to a specific object, can be derived: 1) a set of filtered, overlapping soft masks, and 2) a set of non-overlapping hard masks, which is generated by assigning each pixel to the unique query that yielded the highest probability at its location. The outputs will be utilized to guide object-centric token merging in Sec.~\ref{Object-centric Token Merging}.

\textbf{LLMs as Language Head Decoder.} To bridge our object-centric visual representations with the language modality, we employ a projector layer which is designed as a flexible and configurable MLP. Once projected, the sequence of visual tokens is fed into the LLM, along with the embedded input text prompt. The LLM functions as the final language decoder, auto-regressively generating the textual response by attending to both the compact object-centric visual context and the user's query.

\subsection{Object-centric Token Merging}
\label{Object-centric Token Merging}

\begin{figure} [t]
    \centering
    \includegraphics[width=\linewidth]{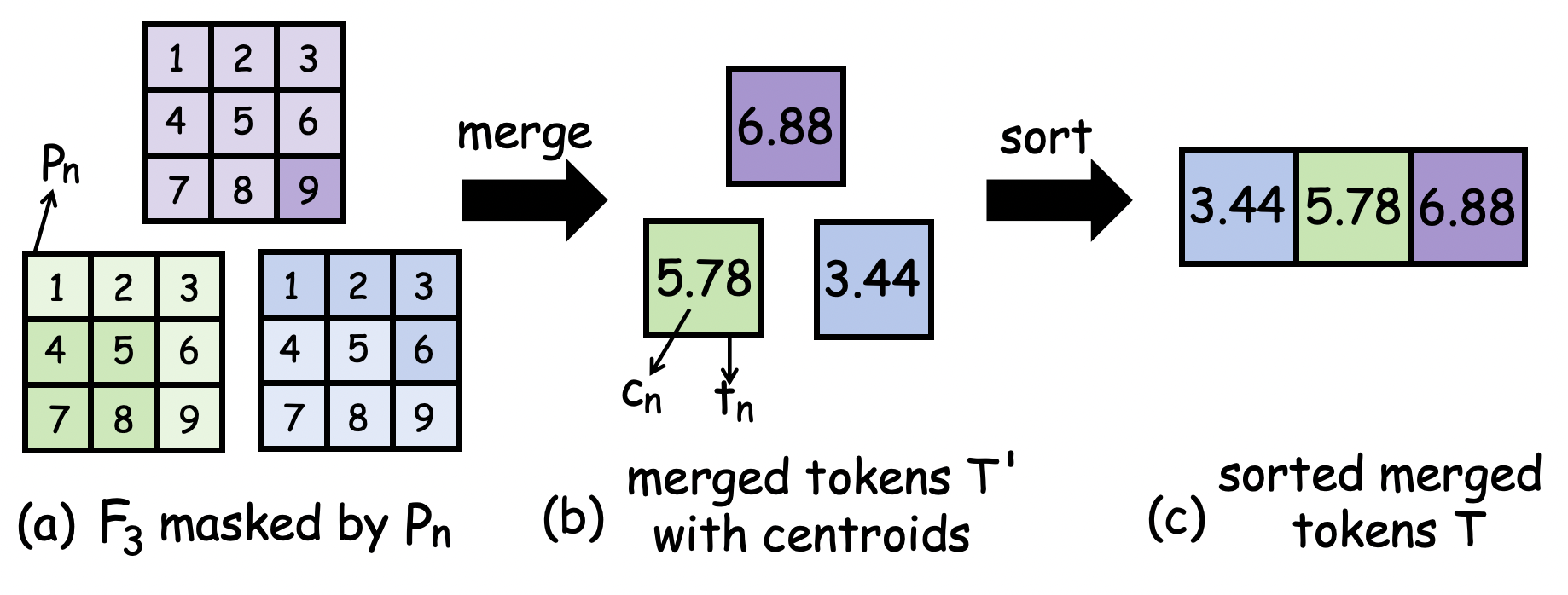}
    \caption{\textbf{Illustration of Centroid-Guided Sorting.} Assume $N=3$. In Step (a), the number in a token indicates the $i$-th token. For simplicity, darker (lighter) tokens represent a 0.9 (0.1) weight in $P_n$. In Step (b), the number in token $t_n$ indicates the centroid position $c_n$. The tokens are merged without sorting. Step (c) shows the final merged tokens $T$ sorted in ascending order based on their centroid values.}
    \label{sort_soft_mask}
\end{figure}

We flatten \(F_3\), which is from $F_C=\{F_0,F_1,F_2,F_3\}$, into \(F\in \mathbb{R}^{H W \times C}\) and let a single token from \(F\) be \(f_i \in \mathbb{R}^C\). We aim to produce a single feature token \(t_n \in \mathbb{R}^C\) for each mask \(P_n \in\mathcal{P}_{\text {valid}}\) by performing a weighted average over the entire visual feature map \(F\). To achieve this, each 2D mask \(P_n\) of shape \(H \times W\) is first flattened into a weight vector \(\Omega_n \in \mathbb{R}^{H W}\), where each element \(\omega_{n, i}\) corresponds to the \(i\)-th token. The index \(i\) follows the raster scanning sequence. This operation is repeated for every valid mask \(P_n\), yielding a set of \(N\) aggregated object tokens \(T'=\left\{t_1, t_2, \ldots, t_N\right\}\). To ensure the final token sequence follows an order consistent with the spatial positions in the image, we then calculate the centroid of each mask and sort the tokens accordingly. The centroid position \(c_n\) of each mask \(P_n\) is similarly computed via the weighted average. The merged object token \(t_n \in \mathbb{R}^C\) and \(c_n\) are formulated as follows:

\begin{equation}
    t_n=\frac{\sum_{i=1}^{H W} \omega_{n, i} \cdot f_i}{\sum_{i=1}^{H W} \omega_{n, i}}
    \qquad 
    c_n=\frac{\sum_{i=1}^{H W} \omega_{n, i} \cdot i}{\sum_{i=1}^{H W} \omega_{n, i}}
\label{merge_equation}
\end{equation}

\noindent where \(\sum_{i=1}^{H W} \omega_{n, i} \cdot f_i\) is a weighted sum, which iterates over all pixel feature vectors \(f_i\) and weights each one according to its value \(\omega_{n, i}\) from the \(n\)-th mask. \(\sum_{i=1}^{H W} \omega_{n, i}\) is the sum of all values in the \(n\)-th mask and serves to normalize the weighted sum of features. Subsequently, the set of merged tokens \(T'\) is sorted in ascending order based on their centroid values \(\left\{c_1, \ldots, c_N\right\}\), yielding a final and spatially-ordered visual representation \(T\), as shown in Fig.~\ref{sort_soft_mask}. The sorted tokens \(T\) are then input into the projector layer. We adopt two merging strategies: \textit{merging via soft masks} and \textit{merging via hard masks}.

\textbf{Merging via Soft Masks.} When CORE uses the soft masks to merge, each soft mask \(P_n\in[0,1]^{H \times W}\) is a probability map and $\omega_{n, i}$ is the probability of each token. The soft-mask-based method leverages the full probability maps from the segmentation head. Its primary advantage is the preservation of fine-grained information, as it naturally handles ambiguous boundaries and regions of object overlap by assigning partial and probabilistic weights.

\textbf{Merging via Hard Masks.} When hard masks are utilized to merge, each hard mask \(P_n\in\{0,1\}^{H \times W}\) is a binary map and $\omega_{n, i}$ takes the value 1 when the hard mask includes the \(i\)-th feature and 0 when the mask excludes the \(i\)-th feature. The merged object token \(t_n\) is computed as the arithmetic mean of the selected features.  Similar to the soft-mask approach, the resulting set of \(N\) object tokens \(T\) is then spatially sorted based on the centroids of their corresponding hard mask regions \(c_n\). This method first assigns each visual token to the single query with the highest probability. This process yields a set of discrete and non-overlapping object regions with crisp boundaries, ensuring unambiguous representations.

\subsection{Training and Inference Strategy}

\textbf{Training.} Our training strategy follows the two-stage paradigm of LLaVA \citep{liu2023llava}, consisting of feature alignment pretraining and visual instruction tuning. During the first stage, the parameters of the vision backbone (ConvNeXt-L), the segmentation head (Mask2Former), and the LLM are all frozen. The vision components operate as a fixed feature extractor to produce compact object-centric tokens. Only the projector layer is trainable. The projector is trained on a large-scale dataset of image-caption pairs using a standard auto-regressive loss \(\mathcal{L}_{\text {text}}\) \citep{liu2023llava,radford2018improving}. During the second stage, in addition to finetuning the visual projector, we use LoRA \citep{hu2022lora} to finetune the LLM. The loss function is still the auto-regressive loss \(\mathcal{L}_{\text {text}}\) but it is trained with high-quality visual instruction dialogue data.

\textbf{Inference.} During the evaluation phase, our model generates responses in an end-to-end manner for a given image-text pair. We employ a deterministic decoding strategy to ensure reproducible and optimal results.

%% file: sec/4_experiments.tex
\section{Experiments}

\subsection{Implementation Details}
\label{Implementation Details}
\textbf{Model Configuration.} Our CORE model is built upon the MMDetection \citep{mmdetection} framework and we conduct all our experiments on a server equipped with 4 NVIDIA A800 GPUs (80GB). Our visual perception module consists of a ConvNeXt-L \citep{liu2022convnet} and a Mask2Former \citep{cheng2021mask2former}, which we keep frozen. The module is initialized with weights pretrained by OMG-Seg \citep{OMGSeg}. To accelerate inference, the visual perception module operates in full precision (FP32) during training and is switched to half precision (FP16) for evaluation. The input image is first resized to a resolution of \(1024\times1024\). The ConvNeXt-L hierarchical vision backbone processes the input image through four stages, producing a feature pyramid \(F_0,F_1,F_2,F_3\) with progressively increasing channel dimensions of 192, 384, 768, and 1536. For Mask2Former's structure, the number of blank queries is set to 300 and the transformer decoder has \(L=9\) layers. Based on pretraining, the model can recognize 80 \textit{thing} and 53 \textit{stuff} categories. Others are regarded as a single \textit{special} category. In subsequent processing, we treat these 134 categories equally. 
The projector is a two-layer MLP with a GELU activation function, mapping the visual features from a dimension of 1536 to the LLM's hidden dimension of 4096. We choose InternLM2-7B \citep{cai2024internlm2} as CORE's language decoder. For a deep and parameter-efficient fine-tuning of the LLM, we utilize the LoRA \citep{hu2022lora} methodology. Specifically, we configure the adapter with a high rank (\(r=512\)), apply regularization via a scaling alpha of 256 and a dropout rate of 0.05, and do not train any bias terms. 

\textbf{Training and Evaluation Dataset.} Our two-stage training process follows the paradigm established by the LLaVA family \citep{liu2023llava, liu2024llavanext}. In the first stage, feature alignment pretraining, we train only the projector using the LLaVA 558K Mixture dataset \citep{liu2023llava}, which consists of 558K image-caption pairs. For the second stage, visual instruction tuning, we finetune both the projector and the LLM's LoRA adapters using the advanced multimodal instruction dataset LLaVA-NeXT data \citep{liu2024llavanext}. For evaluation, we evaluate our CORE model on six different multimodal tasks, including POPE \citep{li2023evaluating}, MME \citep{fu2024mmecomprehensiveevaluationbenchmark}, MMBench-CN \citep{liu2024mmbench}, ScienceQA-IMG \citep{lu2022learn}, SEEDBench-IMG \citep{li2023seed} and MMMU \citep{yue2023mmmu}.

\subsection{Main Results}
\label{Main Results}

\begin{figure*}[t]
\begin{minipage}[valign=T]{0.6\textwidth}
\captionof{table}{\textbf{Comparison on Fixed-rate Compression Tasks.} For fair comparison, the blue percentage values show the retained performance with fixed tokens, compared with full-token CORE model (ConvNeXt-L backbone) which serves as the 100\% baseline.}
\label{Comparison on Fixed-rate Compression Tasks}
\small
\setlength{\tabcolsep}{2pt}
    \begin{tabular}{l|c|cccccc}
    \toprule
         &  tokens&  POPE&  MME&  MMB$^{\text{CN}}$&  SQA$^{\text{I}}$ &  SEED$^{\text{I}}$ &  MMMU\\ \midrule
         LLaVA-NeXT-7B&  2880&  86.8&  1511.8&  57.3&  67.5&  70.2& 35.1\\
         \cellcolor{mycolor!60}\textbf{CORE (vanilla)}&  \cellcolor{mycolor!60}1024&  \cellcolor{mycolor!60}86.4&  \cellcolor{mycolor!60}1626.7&  \cellcolor{mycolor!60}61.0&  \cellcolor{mycolor!60}68.3&  \cellcolor{mycolor!60}69.6& \cellcolor{mycolor!60}36.8\\
 \cellcolor{mycolor!60}& \cellcolor{mycolor!60}& \cellcolor{mycolor!60}\textcolor{blue}{100.0\%}& \cellcolor{mycolor!60}\textcolor{blue}{100.0\%}& \cellcolor{mycolor!60}\textcolor{blue}{100.0\%}& \cellcolor{mycolor!60}\textcolor{blue}{100.0\%}& \cellcolor{mycolor!60}\textcolor{blue}{100.0\%}&\cellcolor{mycolor!60}\textcolor{blue}{100.0\%}\\\midrule
         ToMe \citep{bolya2022token}& 720& 85.3& 1407.8&  55.6& 67.2& --&--\\
         FastV \citep{chen2024image}&  640&  79.5&  1412.6&  53.5&  67.4&  --& --\\
         PDrop \citep{xing2025pyramiddropacceleratinglargevisionlanguage}&  640&  83.8&  1475.9&  55.2&  66.7&  --& --\\
         SparseVLM \citep{zhang2025sparsevlmvisualtokensparsification}&  640&  85.3&  1456.8&  58.6&  67.6&  --& 34.6\\
         PruMerge+ \citep{shang2025prumerge}&  640&  85.3&  1480.2&  57.3&  67.8&  --& --\\
         TRIM \citep{song2024moresimpleeffectivetoken}&  640&  \textbf{86.9}&  1471.8&  55.8&  66.9&  --& --\\
         VisionZip \citep{yang2024visionzip}&  640&  86.0&  1493.4&  58.1&  68.1&  66.7& 34.7\\
         DART \citep{wen2025stop}&  640&  85.0&  1450.2&  57.1&  68.2&  --& --\\
 DivPrune \citep{alvar2025divprune}& 640& \textbf{86.9}& 1469.7& 57.3& 67.8& \textbf{67.6}&36.9\\
 \cellcolor{mycolor!60}\textbf{CORE (ours)}& \cellcolor{mycolor!60}640& \cellcolor{mycolor!60}\textbf{86.9}& \cellcolor{mycolor!60}\textbf{1521.6}& \cellcolor{mycolor!60}\textbf{60.0}& \cellcolor{mycolor!60}\textbf{69.2}& \cellcolor{mycolor!60}\textbf{67.6}&\cellcolor{mycolor!60}\textbf{38.3}\\
 \cellcolor{mycolor!60}& \cellcolor{mycolor!60}& \cellcolor{mycolor!60}\textcolor{blue}{100.6\%}& \cellcolor{mycolor!60}\textcolor{blue}{93.5\%}& \cellcolor{mycolor!60}\textcolor{blue}{98.4\%}& \cellcolor{mycolor!60}\textcolor{blue}{101.3\%}& \cellcolor{mycolor!60}\textcolor{blue}{97.1\%}&\cellcolor{mycolor!60}\textcolor{blue}{104.1\%}\\ \midrule
 ToMe \citep{bolya2022token}& 360& 82.4& 1343.2& 54.6& 67.7& --&--\\
 FastV \citep{chen2024image}& 320& 49.5& 1099.0& 42.5& 66.6& --&--\\
 PDrop \citep{xing2025pyramiddropacceleratinglargevisionlanguage}& 320& 60.8& 1171.5& 44.7& 66.7& --&--\\
 SparseVLM \citep{zhang2025sparsevlmvisualtokensparsification}& 320& 76.9& 1386.1& 56.7& 67.2& --&34.4\\
 PruMerge+ \citep{shang2025prumerge}& 320& 79.5& 1444.3& 55.6& 68.1& --&--\\
 TRIM \citep{song2024moresimpleeffectivetoken}& 320& \textbf{86.5}& 1443.8& 51.0& 66.2& --&--\\
 VisionZip \citep{yang2024visionzip}& 320& 82.3& 1397.1& 55.6& 67.5& 63.4&35.3\\
 DART \citep{wen2025stop}& 320& 81.0& 1419.5& 55.7& 67.5& --&--\\
 DivPrune \citep{alvar2025divprune}& 320& 84.7& 1423.3& 55.7& 67.7& 65.4&37.1\\
  \cellcolor{mycolor!60}\textbf{CORE (ours)}& \cellcolor{mycolor!60}320& \cellcolor{mycolor!60}86.3& \cellcolor{mycolor!60}\textbf{1497.9}& \cellcolor{mycolor!60}\textbf{57.3}& \cellcolor{mycolor!60}\textbf{69.4}& \cellcolor{mycolor!60}\textbf{65.9}&\cellcolor{mycolor!60}\textbf{38.4}\\
 \cellcolor{mycolor!60}& \cellcolor{mycolor!60}& \cellcolor{mycolor!60}\textcolor{blue}{99.9\%}& \cellcolor{mycolor!60}\textcolor{blue}{92.1\%}& \cellcolor{mycolor!60}\textcolor{blue}{93.9\%}& \cellcolor{mycolor!60}\textcolor{blue}{101.6\%}& \cellcolor{mycolor!60}\textcolor{blue}{94.7\%}&\cellcolor{mycolor!60}\textcolor{blue}{104.3\%}\\ \midrule
 ToMe \citep{bolya2022token}& 180& 73.6& 932.9&  34.0& 64.1& --&--\\
 PruMerge+ \citep{shang2025prumerge}& 160& 71.1& 1289.6& 48.9& 66.9& --&--\\
 TRIM \citep{song2024moresimpleeffectivetoken}& 160& 84.8& 1275.8& 45.2& 65.5& --&--\\
 VisionZip \citep{yang2024visionzip}& 160& 74.9& 1327.8& 50.4& 67.9& 58.3&36.1\\
 DART \citep{wen2025stop}& 160& 75.3& 1325.4& 53.6& 67.8& --&--\\
DivPrune \citep{alvar2025divprune}& 160& 80.0& 1356.6& 53.7& 67.1& 62.5&36.4\\
 \cellcolor{mycolor!60}\textbf{CORE (ours)}& \cellcolor{mycolor!60}160& \cellcolor{mycolor!60}\textbf{86.0}& \cellcolor{mycolor!60}\textbf{1405.3}& \cellcolor{mycolor!60}\textbf{56.7}& \cellcolor{mycolor!60}\textbf{69.8}& \cellcolor{mycolor!60}\textbf{64.7}&\cellcolor{mycolor!60}\textbf{36.6}\\
 \cellcolor{mycolor!60}& \cellcolor{mycolor!60}& \cellcolor{mycolor!60}\textcolor{blue}{99.5\%}& \cellcolor{mycolor!60}\textcolor{blue}{86.4\%}& \cellcolor{mycolor!60}\textcolor{blue}{93.0\%}& \cellcolor{mycolor!60}\textcolor{blue}{102.2\%}& \cellcolor{mycolor!60}\textcolor{blue}{93.0\%}&\cellcolor{mycolor!60}\textcolor{blue}{99.5\%}\\ \bottomrule
    \end{tabular}
\end{minipage}%
\hfill
\begin{minipage}[valign=T]{0.38\textwidth}
    \centering
    
    \begin{minipage}{0.48\linewidth}
        \centering
        \includegraphics[width=\linewidth]{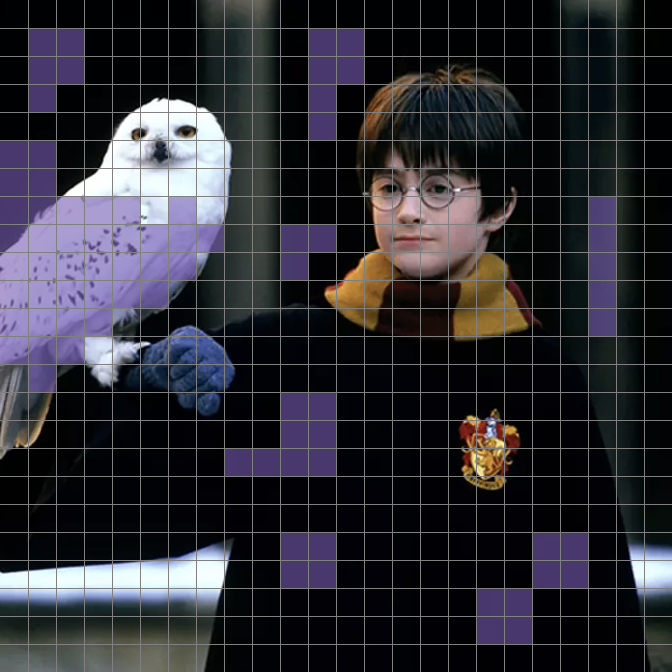}
        \subcaption*{($a_1$) ToMe} 
    \end{minipage}
    \hfill
    \begin{minipage}{0.48\linewidth}
        \centering
        \includegraphics[width=\linewidth]{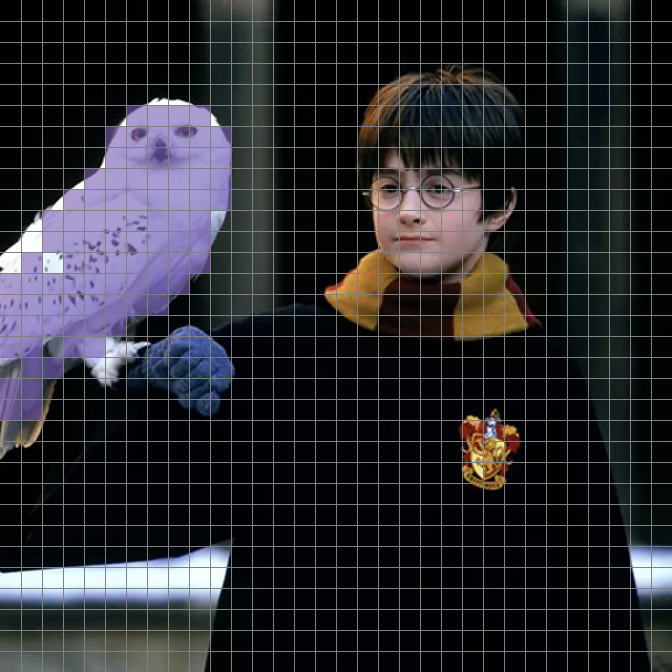}
        \subcaption*{($a_2$) CORE}
    \end{minipage}

    \begin{minipage}{0.48\linewidth}
        \centering
        \includegraphics[width=\linewidth]{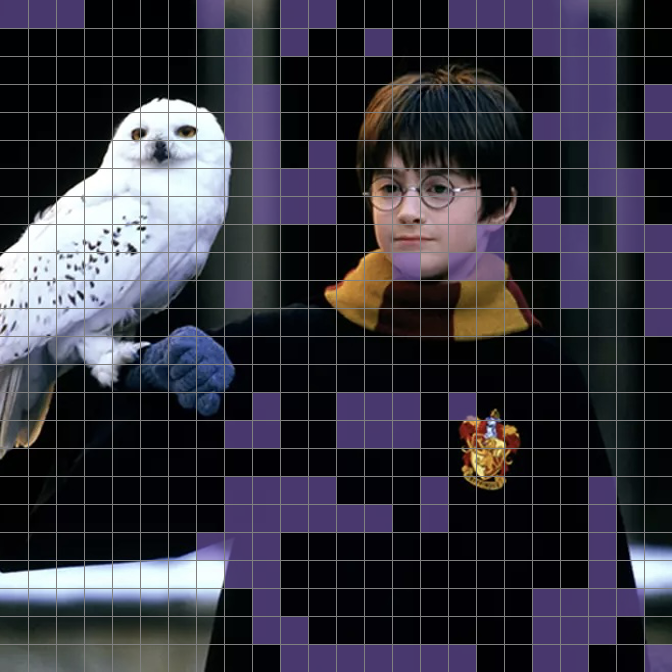}
        \subcaption*{($b_1$) ToMe}
    \end{minipage}%
    \hfill
    \begin{minipage}{0.48\linewidth}
        \centering
        \includegraphics[width=\linewidth]{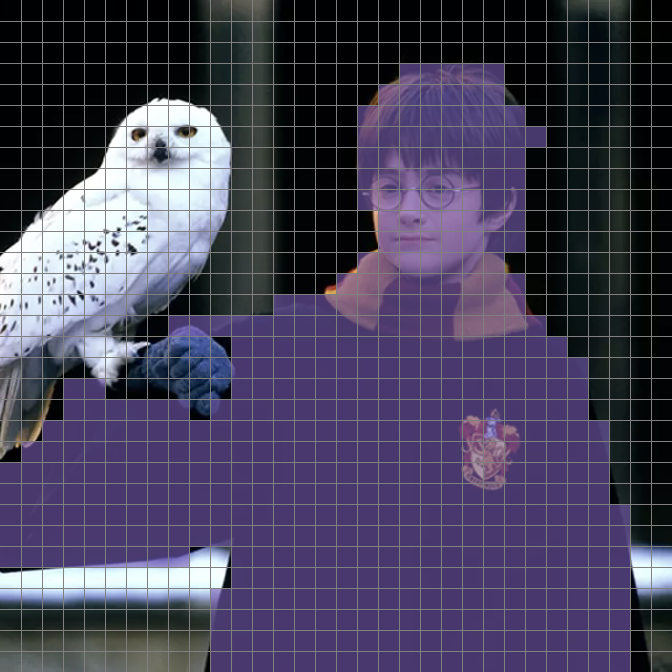}
        \subcaption*{($b_2$) CORE}
    \end{minipage}

    \begin{minipage}{0.48\linewidth}
        \centering
        \includegraphics[width=\linewidth]{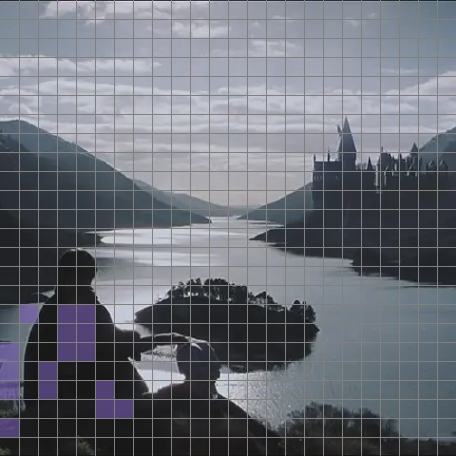} 
        \subcaption*{($c_1$) ToMe}
    \end{minipage}%
    \hfill
    \begin{minipage}{0.48\linewidth}
        \centering
        \includegraphics[width=\linewidth]{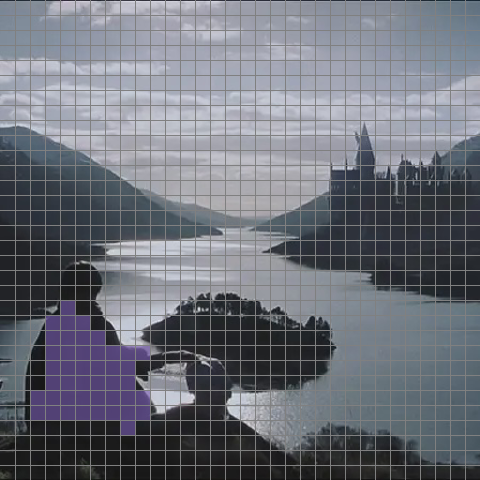} 
        \subcaption*{($c_2$) CORE}
    \end{minipage}

    \begin{minipage}{0.48\linewidth}
        \centering
        \includegraphics[width=\linewidth]{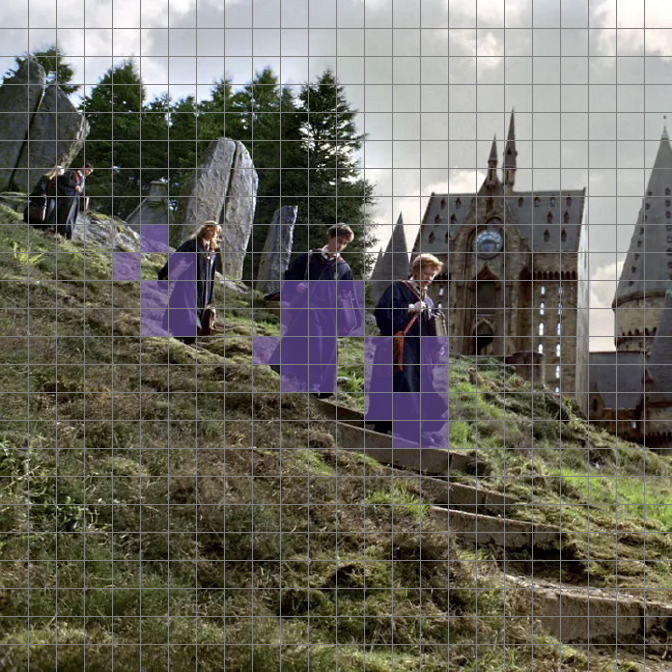} 
        \subcaption*{($d_1$) ToMe}
    \end{minipage}%
    \hfill
    \begin{minipage}{0.48\linewidth}
        \centering
        \includegraphics[width=\linewidth]{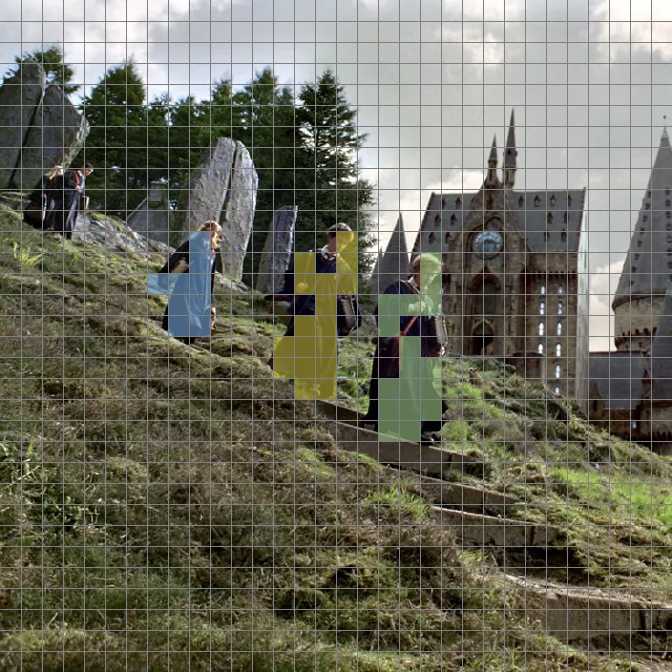} 
        \subcaption*{($d_2$) CORE}
    \end{minipage}
    
    \captionof{figure}{Visualization Comparison}
    \label{Visualization Comparison}

\end{minipage}

\end{figure*}

Under specific circumstances, we also need a fixed-rate token merging strategy which could retain a constant number of tokens to be the input of LLMs. This enables batch processing efficiency and fair performance comparison. Based on the hard masks, we design the fixed-rate strategy by merging tokens in the order of objects. We assume that objects with more tokens (larger objects) have more information redundancy and give priority to merging these objects. Specifically, we first sort the objects in descending order based on their area, which is determined by the number of tokens in each corresponding mask. Subsequently, each object is merged according to the sequence of raster scanning (i.e.,from top to bottom and left to right). This process can be represented by pseudo code Algorithm~\ref{fixed-rate token merging-l} in Sec.~\ref{Algorithms} of Supplementary Material. Besides, for ablation study, we also discuss the small-object-first strategy in Sec.~\ref{ablation} of Supplementary Material.

We compare our CORE model with recent token merging or pruning works based on LLaVA-NeXT in a fixed-rate setting, which can handle high-resolution tasks better. For the ToMe baseline, we select a target token count that is both comparable to other methods and corresponds to a simple fractional keep rate. As shown in Table~\ref{Comparison on Fixed-rate Compression Tasks}\footnotemark, our CORE method achieves state-of-the-art or highly competitive performance across all six tasks. To verify that CORE's high performance is attributable to our compression strategy itself, not to the change in vision encoder, the values in blue indicate the percentage of performance retained by each compressed model relative to the uncompressed, full-token CORE baseline. In some circumstances, our compressed CORE performs better than the full-token baseline. We attribute this to the powerful regularization effect of our object-centric merging strategy. By forcing the model to focus on salient, semantically coherent object representations, it effectively suppresses overfitting to noise and artifacts in the training data, thereby improving the model's generalization capabilities on downstream tasks.

\footnotetext{Some of the results are taken from those reported in \citep{yang2024visionzip,jiang2025visa,dong2025mmtok,zhang2025beyond}. When targeting a 160-token budget, the number of tokens for a few highly complex images slightly exceeded this limit: 4 images from MMB-CN and 8 from SEED-I fall into the [160, 170) token range, while an additional 5 images from SEED-I fall into the [170, 180) token range.}

For dynamic adaptive tasks, Tab.~\ref{Comparison on Dynamic Adaptive Compression Tasks}\footnotemark provides a comprehensive comparison, benchmarking CORE's soft-mask and hard-mask variants against competing object-centric VLMs that utilize methods such as Slot Attention \citep{locatello2020object}. The experimental results demonstrate that the hard-mask-based variant of CORE consistently outperforms all other methods.

\subsection{Visualization Comparison}

\begin{table}
    \centering
\caption{\textbf{Comparison on Dynamic Adaptive Compression Tasks.} Bold indicates the best performance on each dataset.}
\label{Comparison on Dynamic Adaptive Compression Tasks}
\small
\setlength{\tabcolsep}{0.05pt}
    \begin{tabular}{l|cccccc}
    \toprule
         &  POPE&  MME&  MMB$^{\text{CN}}$&  SQA$^{\text{I}}$&  SEED$^{\text{I}}$& MMMU\\ \midrule
         SEED-LLaMA \citep{ge2023making}&  78.0&  1123.9&  --&  --&  48.6& 26.8\\
         Slot-MLLM-base \citep{chi2025slotmllmobjectcentricvisualtokenization}&  78.3&  1128.4&  --&  --&  44.7& 29.0\\
         Slot-MLLM \citep{chi2025slotmllmobjectcentricvisualtokenization}&  79.8&  1202.6&  --&  --&  47.4& 28.0\\ \midrule
         CORE (soft mask)&  83.6&  1339.1&  53.6
&  69.0&  60.3& 37.0\\
         CORE (hard mask)&  \textbf{85.6}&  \textbf{1396.7}
&  \textbf{55.3}&  \textbf{69.9}
&  \textbf{63.1}& \textbf{38.7}\\ \bottomrule
    \end{tabular}  
\end{table}
\afterpage{\footnotetext{Some of the results are taken from those reported in \citep{chi2025slotmllmobjectcentricvisualtokenization}.}}

In this section, we compare our merging strategy with ToMe~\citep{bolya2022token}, as shown in Fig.~\ref{Visualization Comparison}. We ensure that ToMe and CORE have the same compression ratio in each group of pictures and visualize which tokens are planned to be merged. Besides erroneous merging of semantically distinct objects, layer-by-layer merging based on feature similarity is also susceptible causing the ``boundary bleeding'' problem. Fig.~\ref{Visualization Comparison}($a_1$) shows that more of the black background is further integrated, as the boundary of the white owl is mixed with a black background. Besides, as Fig.~\ref{Visualization Comparison}($b_1$) shows, when the character's clothing is visually similar to the background, incorrect merging is likely to happen. Our CORE model fundamentally prevents these issues by leveraging segmentation masks, as Fig.~\ref{Visualization Comparison}($a_2$) and Fig.~\ref{Visualization Comparison}($b_2$) demonstrate. By comparing Fig.~\ref{Visualization Comparison}($c_1$) and ($c_2$), we can see CORE maintains stable segmentation performance in a dim environment. In Fig.~\ref{Visualization Comparison}($d_1$), similar characters are merged together without distinguishing between them by ToMe. However, CORE can distinguish different objects of the same class. (Different mask colors represent different token groups.) These visualizations provide compelling evidence that by shifting the merging criterion from fragile feature affinity to robust semantic identity, CORE produces qualitatively superior and more reliable object-centric representations.

\begin{figure}[t]
    \centering 

    \begin{minipage}[b]{0.20\textwidth}
        \centering
        \includegraphics[width=\linewidth]{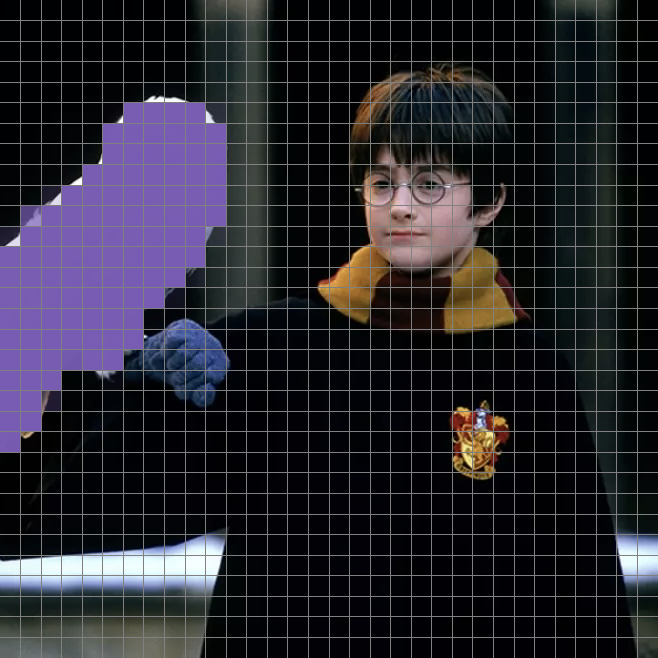}
        \subcaption{the 14th Soft Mask}
        \label{soft_0}
    \end{minipage}
    \begin{minipage}[b]{0.20\textwidth}
        \centering
        \includegraphics[width=\linewidth]{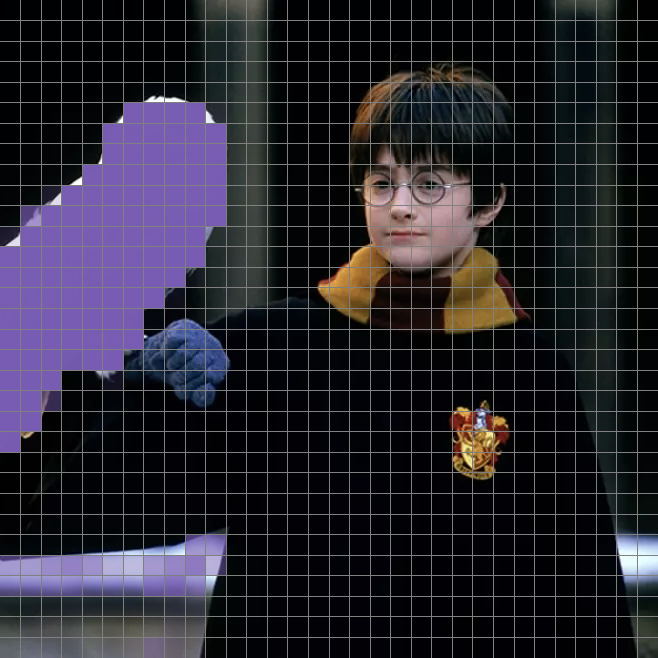} 
        \subcaption{the 18th Soft Mask}
        \label{soft_1}
    \end{minipage}

    \caption{CORE Visualization (soft mask)}
    \label{CORE Visualization (soft mask)}
\end{figure}

On the other hand, Tab.~\ref{Comparison on Dynamic Adaptive Compression Tasks} demonstrates the superiority of hard-mask-guided merging over its soft-mask alternative.  To understand this performance gap, we also visualize the complete soft masks in Sec.~\ref{Soft masks Visualization} in Supplementary Material, representing their weights with varying color intensities. As shown in Fig.~\ref{CORE Visualization (soft mask)}, the same owl object is represented by both the 14th and 18th soft masks. This redundancy introduces ambiguity for the LLM, leading to potential errors in tasks such as object counting and understanding spatial relationships. In contrast, hard masks are mutually exclusive, which enforces a strict ``one object, one token'' mapping and avoids such confusion.

\subsection{Efficiency Analysis}

\begin{table}
    \centering
    \caption{\textbf{Computational Cost and Parameter Comparison of Different Visual Modules.} We calculate the total FLOPs and parameter count for the visual module of CORE.}
    \label{Computational Cost and Parameter Comparison of Different Visual Modules}
    \small
    \begin{tblr}{
        colspec={c|c|cc}, 
        colsep = 3pt,
        hline{1} = {solid}, 
        hline{2} = {solid}, 
        hline{3} = {solid}, 
        hline{5} = {solid}, 
        hline{6} = {solid} 
    }
         &Visual Module&  FLOPs$\downarrow$& Parameters$\downarrow$\\
         LLaVA-NeXT& CLIP ViT-L/14&1.91T  & 303.2 M\\
         \SetCell[r=3]{c}{CORE} & \makecell{ConvNeXt-L \\ Backbone} & 1.44T  & 199.8 M\\ 
         & \makecell{Mask2Former \\Head} &  0.30T &  37.3 M\\ 
         & \textcolor{blue}{Sum}& \textbf{1.74T}&\textbf{237.1M}\\
    \end{tblr}
\end{table}

Our proposed CORE features an enhanced vision module based on LLaVA-NeXT. Tab.~\ref{Computational Cost and Parameter Comparison of Different Visual Modules} compares their computational costs and parameters, showing that our vision module's burden is comparable to, or even slightly lower than, that of the vision encoder in LLaVA-NeXT. We also benchmark the efficiency of CORE against LLaVA-NeXT-7B and competing token compression methods on fixed-rate tasks by measuring the total time to run inference over the full POPE dataset on a single A800 GPU. The results, summarized in Table~\ref{Efficiency Analysis on Fixed-rate Compression Tasks}\footnotemark, demonstrate that CORE simultaneously achieves the highest performance and the lowest runtime.

\footnotetext{Some of the results are taken from those reported in \citep{yang2024visionzip}.}

\begin{table}
    \centering
\setlength{\tabcolsep}{3pt}
\caption{Efficiency Analysis on Fixed-rate Compression Tasks}
\label{Efficiency Analysis on Fixed-rate Compression Tasks}
\small
    \begin{tabular}{l|c|cc|c}
         \toprule
         &  tokens&  \makecell[l]{Total\\ Time $\downarrow$}&  $\Delta \uparrow$& Score$ \uparrow$\\  \midrule
         LLaVA-NeXT-7B&  2880&  2293s&  --& 86.8\\ \midrule
         FastV \citep{chen2024image} &  160&  1792s&  1.28×& 66.5\\
         SparseVLM \citep{zhang2025sparsevlmvisualtokensparsification} &  160&  1895s&  1.21×& 76.6\\
         CORE&  160&  \textbf{1122s}&  \textbf{2.04×}& \textbf{86.0}\\  \bottomrule
    \end{tabular}

\end{table}

In the case of adaptive compression, we evaluate the efficiency of our CORE model on the POPE dataset under two settings: standard half-precision (FP16) inference and a version further optimized with 4-bit quantization, as shown in Tab.~\ref{Efficiency Analysis on Adaptive Compression Tasks}. CORE's aggressive token reduction, from 2880 down to an average of 63.1, leads to dramatic savings in FLOPs and KV Cache. However, since the 7B model's weights constitute a fixed $\sim$14GB overhead, the reduction in total GPU memory is less substantial. This is addressed by applying 4-bit quantization to the LLM, which cuts the GPU memory by more than one third while incurring only a negligible performance loss.

\begin{table}
    \centering
\caption{Efficiency Analysis on Adaptive Compression Tasks}
\label{Efficiency Analysis on Adaptive Compression Tasks}
\small
\setlength{\tabcolsep}{1pt}
    \begin{tabular}{c|c|cccc} \toprule
         &  \makecell[l]{POPE \\tokens}&  \makecell{FLOPs \\$\downarrow$} &  \makecell[l]{KV \\Cache $\downarrow$}&  \makecell[l]{GPU \\Memory$\downarrow$}& \makecell[l]{POPE\\ Score$\uparrow$}\\ \midrule
         \makecell{LLaVA-NeXT\\-7B} &  2880&  41.7T&  1440.0MB&  16.7GB& 86.8\\  \midrule
         \makecell{CORE\\-FP16}&  63.1&  2.6T&  7.9MB&  15.1GB& 85.9\\ \midrule
         \makecell{CORE\\-4bit}&  63.1&  2.6T&  7.9MB&  5.5 GB& 85.6\\ \bottomrule
    \end{tabular}
       
\end{table}

\subsection{Analysis of Token Counts on Various Datasets}

\begin{figure}
    \centering
    \includegraphics[width=\linewidth]{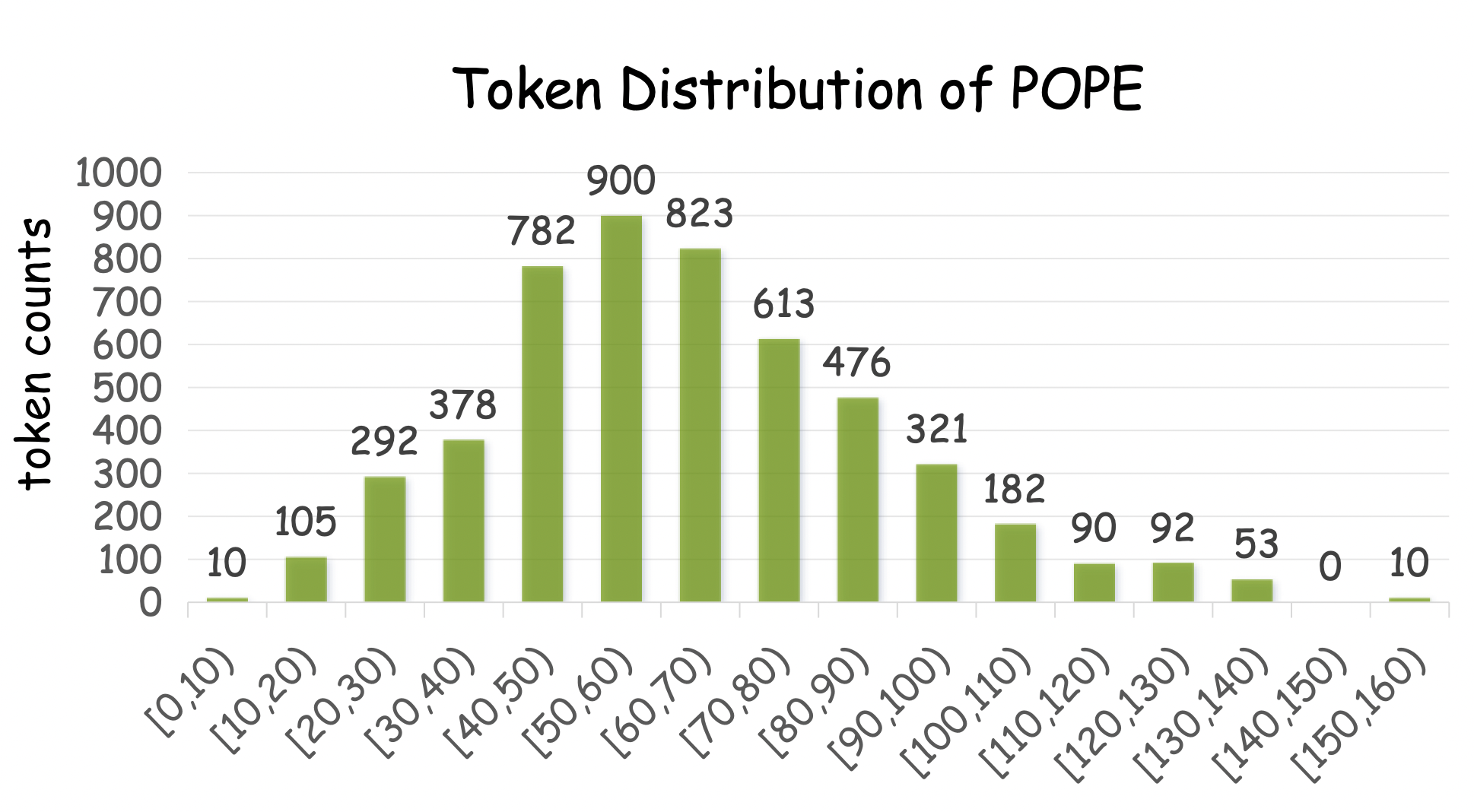}
    \caption{Token Distribution of POPE}
    \label{token_counts_pope1}
\end{figure}

In order to further analyze our adaptive compression strategy, we quantify its compression ability on a per-dataset basis. Tab.~\ref{Adaptive Token Counts on Various Datasets} summarizes these statistics, including the average number of tokens CORE generates for each benchmark. To provide a more granular analysis of our adaptive compression, Fig.~\ref{token_counts_pope1} visualizes the distribution of the number of generated tokens for all images in the POPE dataset using a histogram. Other datasets' visualization can be found in Sec.~\ref{Token Distribution of Datasets} in Supplementary Material.

\begin{table}
    \centering
\caption{Adaptive Token Counts on Various Datasets}
\label{Adaptive Token Counts on Various Datasets}
\small
    \begin{tabular}{l|ccccc} \toprule
         Dataset&  Min&  Max&  Mean  &  Median& SD\\ \midrule
         POPE&  7&  155&  63.1&  61& 25.0\\
         MME&  2&  133&  48.3&  46& 24.1\\
         MMB$^{\text{CN}}$&  2&  162&  35.9&  31& 23.1\\
         SQA$^{\text{I}}$&  3&  104&  25.3&  23& 13.6\\
         SEED$^{\text{I}}$&  6&  179&  61.0&  58& 26.7\\
         MMMU&  1&  140&  22.3&  18& 19.1\\ \bottomrule
    \end{tabular}

\end{table}

\subsection{Discussion on Segmentation Dependency}
\begin{figure}[t]
    \centering 

    \begin{minipage}[b]{0.20\textwidth}
        \centering
        \includegraphics[width=\linewidth]{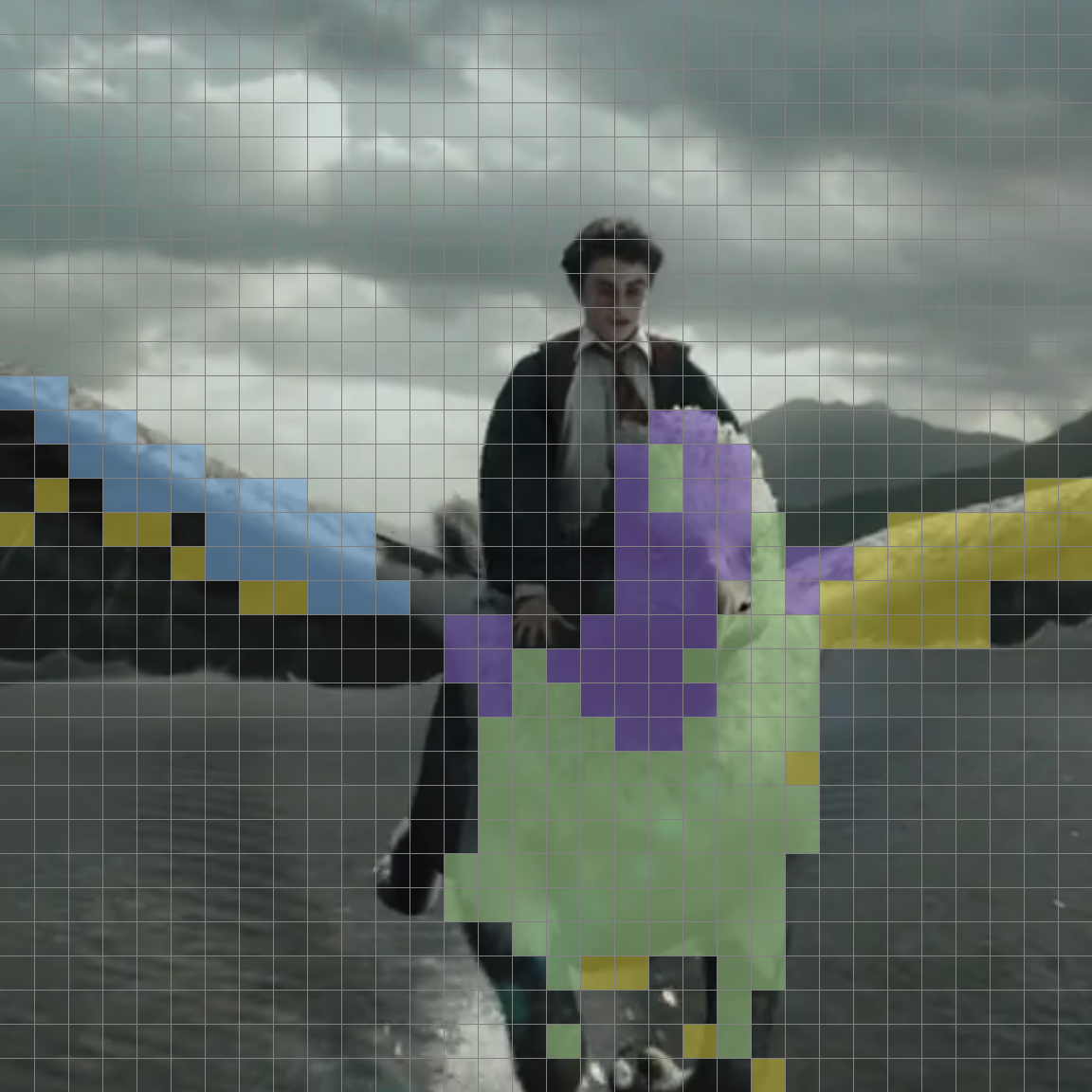}
        \subcaption{Out-of-Distribution}
        \label{problem_0}
    \end{minipage}
    \begin{minipage}[b]{0.20\textwidth}
        \centering
        \includegraphics[width=\linewidth]{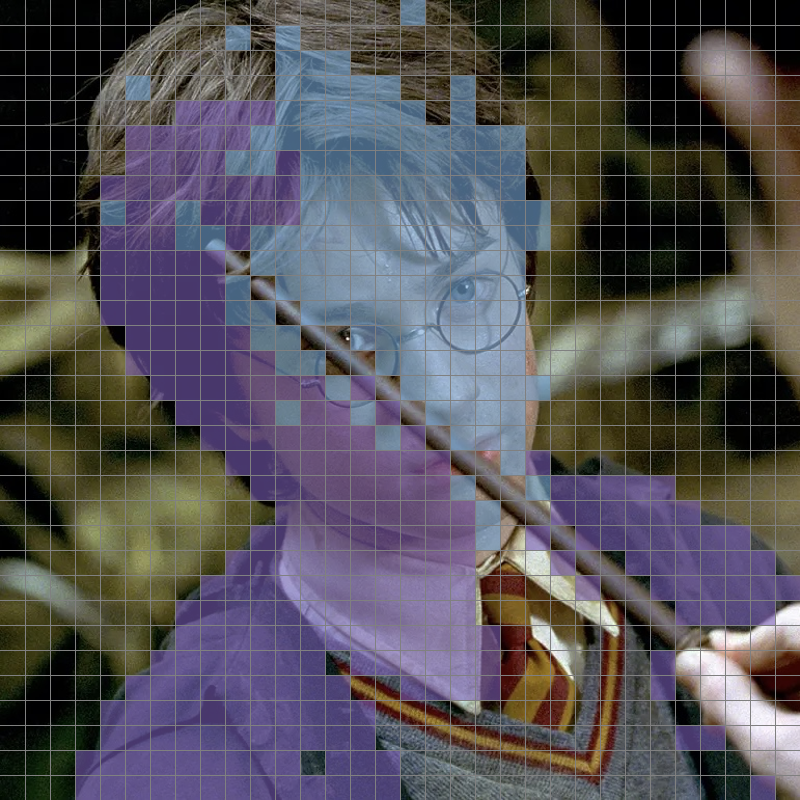} 
        \subcaption{Occluded}
        \label{problem_1}
    \end{minipage}

    \caption{Discussion on Segmentation Dependency}
    \label{Discussion on Segmentation Dependency}
\end{figure}

As Sec.~\ref{Implementation Details} mentions, the Mask2Former segmentation head can only recognize 133 predefined categories. This subsection therefore investigates the robustness of CORE when faced with out-of-distribution (OOD) or occluded objects. The principle behind CORE's object recognition is to use blank queries to actively search for pretrained objects within the image. Fig.~\ref{problem_0} displays an OOD category, specifically a mythical creature. While CORE fails to recognize the entire object, it can still partition the unknown object into several parts based on the features of known objects, successfully completing the merging task. In fact, when the segmentation prior fails to recognize an OOD object, CORE tends to avoid erroneous semantic merging and preserve more tokens. This conservative strategy ensures the integrity of critical visual information. Similarly, Fig.~\ref{problem_1} demonstrates that when an object encounters occlusion, CORE can still complete the recognition task by segmenting it into two parts. However, even under this conservative token-retention mechanism, the maximum token count for our adaptive approach stays below 180 on all benchmarks with 30000+ images, as shown in Sec.~\ref{Token Distribution of Datasets}. This proves that CORE effectively balances robustness with high efficiency.

%% file: sec/5_conclusion.tex
\section{Limitation}
While CORE significantly reduces the theoretical computational load (FLOPs), the complex computational flow and data scheduling within its vision module create a memory bandwidth bottleneck, limiting the model from realizing its full potential in terms of practical inference speed. Future work will address this from a low-level systems optimization perspective, employing techniques such as operator fusion, customized CUDA kernels, and I/O-aware scheduling to further unleash CORE's performance advantages. 

\section{Conclusion}

In this paper, we introduced CORE, a new object-centric paradigm for visual token compression to address the high computational overhead of LVLMs. CORE leverages internally generated segmentation masks as a high-level semantic prior to merge tokens into compact object-level representations, preserving crucial spatial information via centroid-guided sorting. On both fixed-rate and adaptive-rate compression tasks, CORE achieved SOTA performance across multiple datasets. CORE preserves object-level semantic and spatial information, which gives it immense application potential in various fields. These include intelligent image/video retrieval and moderation, environmental perception for robotics and autonomous systems, as well as large-scale video surveillance analysis.

%% file: sec/X_suppl.tex
\makeatletter
\def\maketitlesupplementary
    {
    \twocolumn 
    \begin{center}
        \Large\bf \thetitle
    \end{center}
    \begin{center}
        \vspace{0.5em} 
        \large Supplementary Material
    \end{center}
    \vspace{1.0em} 
    }
\makeatother

\raggedbottom

\setcounter{page}{1}
\maketitlesupplementary

\section{Ablation Study}
\label{ablation}

In Sec.~\ref{Main Results}, we introduce our primary merging heuristic which prioritizes large objects in the case of fixed-rate compression. To investigate the model's sensitivity to this choice, we conduct an ablation study with an inverted \textit{small-object-first} strategy. The results, presented in Tab.~\ref{ablation study table}, show that reversing the order leads to only a minimal drop in average performance. This not only demonstrates CORE's strong robustness to the merging order but also validates our \textit{large-object-first} heuristic as a slightly superior choice. For completeness, the pseudo code for this inverted strategy is detailed as Algorithm~\ref{fixed-rate token merging-s} in Sec.~\ref{Algorithms}.

\begin{table} [h]
    \centering
\caption{\textbf{Ablation Study of Small-object-first Strategy.} Red (Blue) font indicates the performance drop (gain) relative to the large-object-first merging strategy.}
\label{ablation study table}
\small
\setlength{\tabcolsep}{2pt}
    \begin{tabular}{c|ccccccc}
\toprule
 tokens& POPE& MME& MMB$^{\text{CN}}$& SQA$^{\text{I}}$& SEED$^{\text{I}}$& MMMU&Avg.\\  \midrule
         640&  85.3&  1509.3&  59.5&  70.5
&  66.1&  37.3
&  65.7\\
 & \textcolor{red}{-1.6}& \textcolor{red}{-12.3}& \textcolor{red}{-0.5}& \textcolor{blue}{+1.3}& \textcolor{red}{-1.5}& \textcolor{red}{-1.0}&\textcolor{red}{-0.7}\\ \midrule
         320&  86.4&  1486.5
&  58.0
&  69.8
&  64.6&  38.2&  65.2\\
 & \textcolor{blue}{+0.1}& \textcolor{red}{-11.4}& \textcolor{blue}{+0.7}& \textcolor{blue}{+0.4}& \textcolor{red}{-1.3}& \textcolor{red}{-0.2}&\textcolor{red}{-0.2}\\ \midrule
         160&  85.7&  1378.0&  56.5&  70.2&  63.6&  38.6&  63.9\\

 & \textcolor{red}{-0.3}& \textcolor{red}{-27.3}& \textcolor{red}{-0.2}& \textcolor{blue}{+0.4}& \textcolor{red}{-1.1}& \textcolor{blue}{+2.0}&\textcolor{red}{-0.1}\\ \bottomrule
    \end{tabular}     
\end{table}

\section{Algorithms}
\label{Algorithms}

The Algorithm~\ref{fixed-rate token merging-l} first calculates the area of all segmentation regions (the number of tokens contained) and sorts them in descending order by area. Subsequently, it sets a merging budget $\Delta$ which means the total number of tokens to be reduced, based on the difference between the original token count and the target number. During the merging phase, the algorithm iterates through these regions in descending order of size. As long as the budget $\Delta$ is sufficient, it fuses all tokens within a region into a single token via averaging and deducts the corresponding cost ($A_n - 1$) from the budget. If the budget is insufficient to merge the entire current region, the algorithm performs a partial merging to exhaust the remaining budget. Once the budget reaches zero, all tokens from the remaining regions, typically smaller objects are kept intact without merging. Finally, the algorithm returns the token set composed of newly merged tokens and preserved original tokens, totaling $N_{\text{target}}$. Algorithm~\ref{fixed-rate token merging-s} is similar to Algorithm~\ref{fixed-rate token merging-l} but changes the sorting criterion in line 7, prioritizing the merging of small objects.


\begin{algorithm}[H]
\caption{\textbf{Fixed-rate Token Merging.} Larger objects are merged earlier.}
\label{fixed-rate token merging-l}
\begin{algorithmic}[1]
\Require 
    Set of $N$ hard masks $\mathcal{Q}_{\text{valid}} = \{Q_1, \ldots, Q_N\}$,
    Vision features $F \in \mathbb{R}^{HW \times C}$, 
    Target token number $N_{\text{target}}$
\Ensure 
    Merged visual tokens $F_{\text{merged}} \in \mathbb{R}^{N_{\text{target}} \times C}$

\State // \textit{1. Analyze and Prioritize Segments}
\State $\mathcal{S} \leftarrow \text{EmptyList}$ 
\For{$n=1$ to $N$}
    \State $A_n \leftarrow \text{Area}(Q_n)$
    \State Add tuple $(n, A_n, Q_n)$ to $\mathcal{S}$
\EndFor
\State Sort segments $\mathcal{S}$ primarily by descending area $A_n$, then by ascending mask ID $n$

\State // \textit{2. Perform Budgeted Merging}
\State $\Delta \leftarrow |F| - N_{\text{target}}$
\State // \textit{Initialize merging budget (number of tokens to remove)}
\State $F_{\text{merged}} \leftarrow \text{EmptyList}$ 
\For{each segment $(n, A_n, Q_n)$ in sorted $\mathcal{S}$}
    \State $F_n \leftarrow \text{SelectFeatures}(F, Q_n)$  
    \If{$\Delta > 0$ and $(A_n - 1) \le \Delta$} 
        \State // \textit{Case 1: Fully merge the mask}  
        \State $\bar{f}_n \leftarrow \text{AverageFeatures}(F_n)$
        \State $d_n \leftarrow \text{Centroid}(Q_n)$ 
        \State Add token $(\bar{f}_n, d_n)$ to $F_{\text{merged}}$
        \State $\Delta \leftarrow \Delta - (A_n - 1)$
    \ElsIf{$\Delta > 0$ and $(A_n - 1) > \Delta$} 
        \State // \textit{Case 2: Partially merge the mask}  
        \State $\bar{f} \leftarrow \text{AverageFeatures}(\text{the first}\  (\Delta+1) \  \text{tokens of}\  F_n)$
        \State $d \leftarrow \text{Centroid}(\text{the first}\  (\Delta+1) \  \text{tokens of}\  Q_n)$
        \State Add token $(\bar{f}, d)$ to $F_{\text{merged}}$
        \State Add the remaining $A_n-(\Delta+1)$ tokens from $F_n$ to $F_{\text{merged}}$
        \State $\Delta \leftarrow 0$
    \Else 
        \State // \textit{Case 3: No budget left, keep all tokens}
        \State Add all tokens from $F_n$ (with their original positions) to $F_{\text{merged}}$
    \EndIf
\EndFor

\State // \textit{3. Finalize Output}
\State Sort $F_{\text{merged}}$ by spatial position (original or averaged centroid)
\State $F_{\text{merged}} \leftarrow \text{StackFeatures}(F_{\text{merged}})$
\State \Return $F_{\text{merged}}$
\end{algorithmic}
\end{algorithm}

\begin{algorithm}[H]
\caption{\textbf{Fixed-rate Token Merging.} Smaller objects are merged earlier.}
\label{fixed-rate token merging-s}
\begin{algorithmic}[1]
\Require 
    Set of $N$ hard masks $\mathcal{Q}_{\text{valid}} = \{Q_1, \ldots, Q_N\}$,
    Vision features $F \in \mathbb{R}^{HW \times C}$, 
    Target token number $N_{\text{target}}$
\Ensure 
    Merged visual tokens $F_{\text{merged}} \in \mathbb{R}^{N_{\text{target}} \times C}$

\State // \textit{1. Analyze and Prioritize Segments}
\State $\mathcal{S} \leftarrow \text{EmptyList}$ 
\For{$n=1$ to $N$}
    \State $A_n \leftarrow \text{Area}(Q_n)$
    \State Add tuple $(n, A_n, Q_n)$ to $\mathcal{S}$
\EndFor
\State Sort segments $\mathcal{S}$ primarily by \textcolor{red}{ascending} area $A_n$, then by ascending mask ID $n$

\State // \textit{2. Perform Budgeted Merging}
\State $\Delta \leftarrow |F| - N_{\text{target}}$
\State // \textit{Initialize merging budget (number of tokens to remove)}
\State $F_{\text{merged}} \leftarrow \text{EmptyList}$ 
\For{each segment $(n, A_n, Q_n)$ in sorted $\mathcal{S}$}
    \State $F_n \leftarrow \text{SelectFeatures}(F, Q_n)$  
    \If{$\Delta > 0$ and $(A_n - 1) \le \Delta$} 
        \State // \textit{Case 1: Fully merge the mask}  
        \State $\bar{f}_n \leftarrow \text{AverageFeatures}(F_n)$
        \State $d_n \leftarrow \text{Centroid}(Q_n)$ 
        \State Add token $(\bar{f}_n, d_n)$ to $F_{\text{merged}}$
        \State $\Delta \leftarrow \Delta - (A_n - 1)$
    \ElsIf{$\Delta > 0$ and $(A_n - 1) > \Delta$} 
        \State // \textit{Case 2: Partially merge the mask}  
        \State $\bar{f} \leftarrow \text{AverageFeatures}(\text{the first}\  (\Delta+1) \  \text{tokens of}\  F_n)$
        \State $d \leftarrow \text{Centroid}(\text{the first}\  (\Delta+1) \  \text{tokens of}\  Q_n)$
        \State Add token $(\bar{f}, d)$ to $F_{\text{merged}}$
        \State Add the remaining $A_n-(\Delta+1)$ tokens from $F_n$ to $F_{\text{merged}}$
        \State $\Delta \leftarrow 0$
    \Else 
        \State // \textit{Case 3: No budget left, keep all tokens}
        \State Add all tokens from $F_n$ (with their original positions) to $F_{\text{merged}}$
    \EndIf
\EndFor

\State // \textit{3. Finalize Output}
\State Sort $F_{\text{merged}}$ by spatial position (original or averaged centroid)
\State $F_{\text{merged}} \leftarrow \text{StackFeatures}(F_{\text{merged}})$
\State \Return $F_{\text{merged}}$
\end{algorithmic}
\end{algorithm}

\section{Soft masks Visualization}
\label{Soft masks Visualization}

Presented below are the complete set of soft masks for the sample image.

\begin{center}
    \vspace{1em}

    \newcounter{colcounter}\setcounter{colcounter}{0}%

    \foreach \i in {0,...,32}
    {
        \begin{minipage}[b]{0.19\linewidth}
            \centering
            \includegraphics[width=\linewidth]{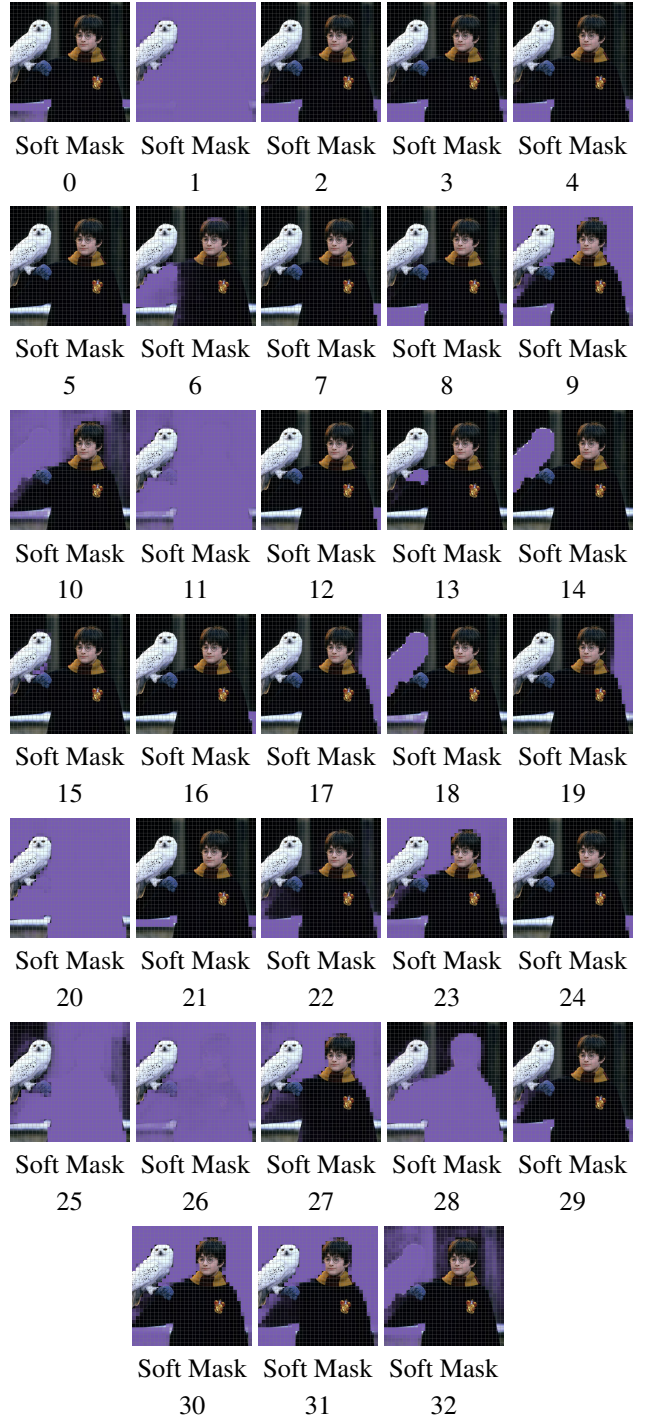}
            \vspace{2pt}
            Soft Mask \i
        \end{minipage}  %
        %
        \stepcounter{colcounter}%
        \ifnum\value{colcounter}=5%
            \par 
            \vspace{0.5em}%
            \setcounter{colcounter}{0}%
        \fi%
    }
    \vspace{1em} 
    \captionof{figure}{\textbf{All Soft Masks.} We can see one object may have multiple soft masks.}
    \label{All Soft Masks} 
\end{center}

\section{Token Distribution of Datasets}
\label{Token Distribution of Datasets}

This section visually illustrates the detailed distribution of the number of tokens generated by the CORE model across multiple datasets when using its adaptive compression strategy. These plots, which use token count intervals as the x-axis and image frequency as the y-axis, consistently exhibit a clear unimodal distribution. This indicates that the vast majority of images are compressed into a relatively concentrated range. This section demonstrates CORE's ability to dynamically adjust its compression rate based on the semantic complexity of the image content or the number of objects, rather than relying on a fixed token count.

\begin{center}
    \begin{minipage}{\linewidth} 
        \centering
        \includegraphics[width=\linewidth]{token_counts_pope.jpg}
        \centerline{(a) Token count analysis on the POPE benchmark.}
    \end{minipage}
    \vspace{1em} 
    
    \begin{minipage}{\linewidth}
        \centering
        \includegraphics[width=\linewidth]{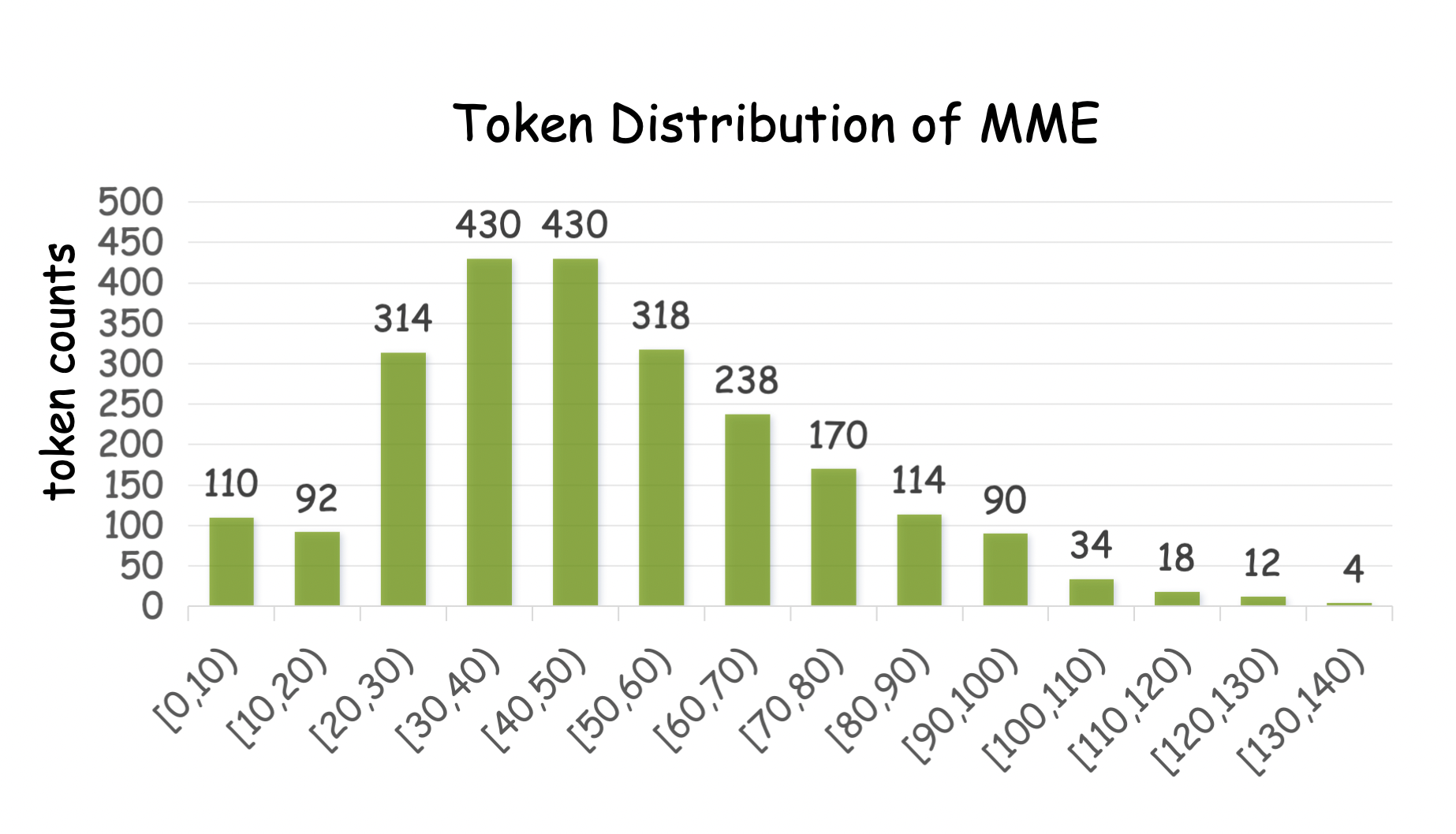}
        \centerline{(b) Token count analysis on the MME benchmark.}
    \end{minipage}
    \vspace{1em}

    \begin{minipage}{\linewidth}
        \centering
        \includegraphics[width=\linewidth]{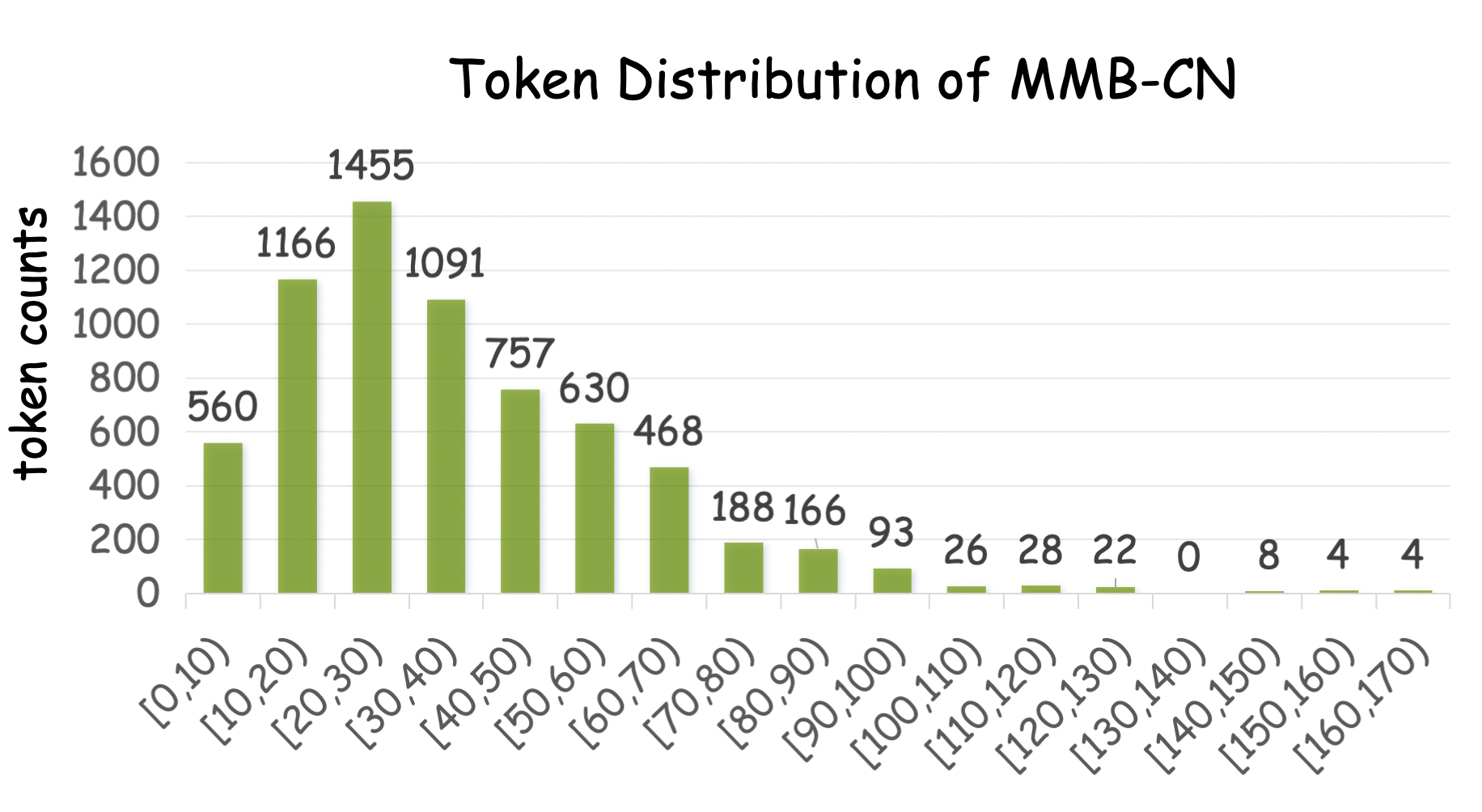}
        \centerline{(c) Token count analysis on the MMB-CN benchmark.}
    \end{minipage}
    \vspace{1em}

    \begin{minipage}{\linewidth}
        \centering
        \includegraphics[width=\linewidth]{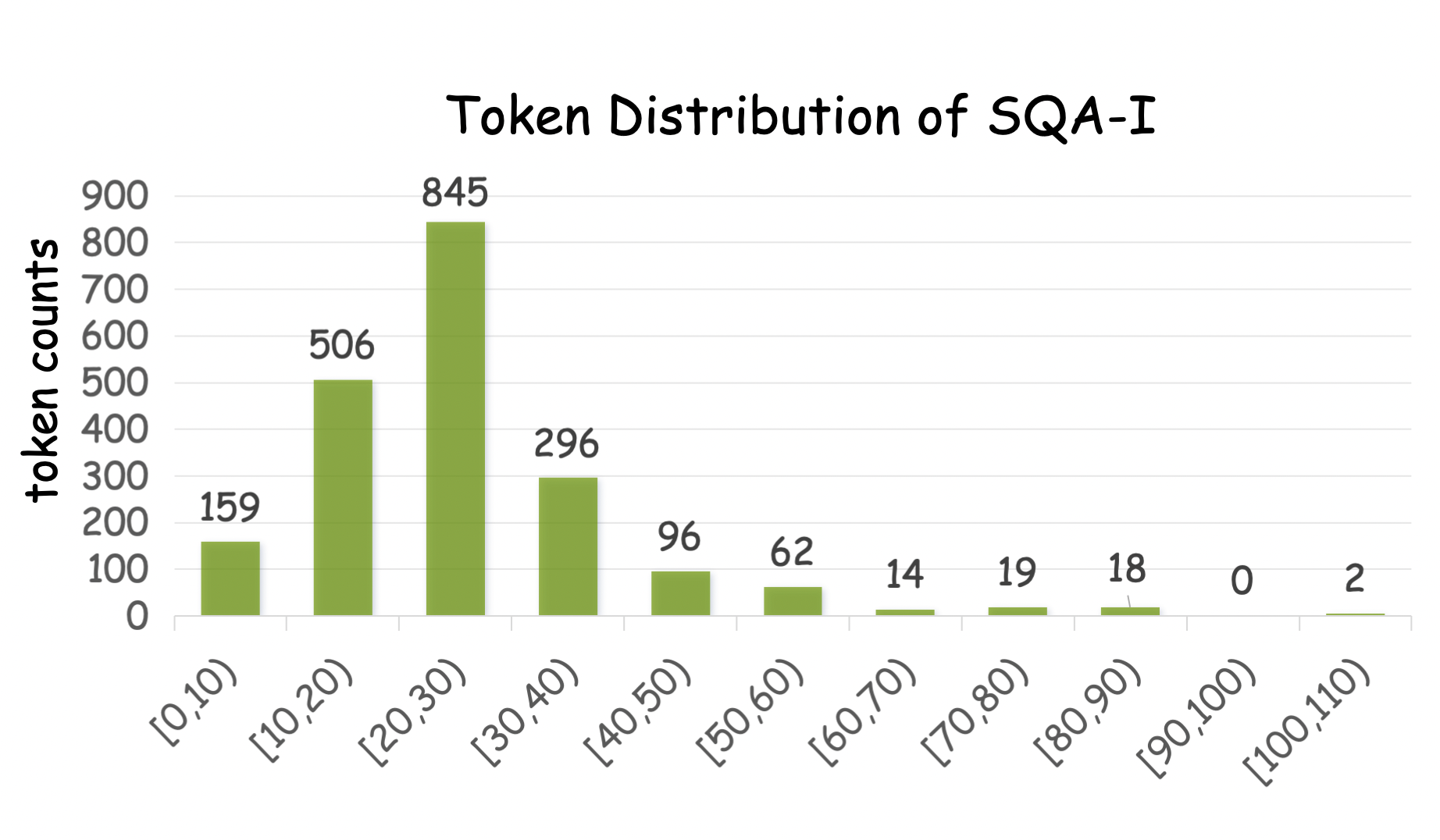}
        \centerline{(d) Token count analysis on the SQA-I benchmark.}
    \end{minipage}
    \vspace{1em}

    \begin{minipage}{\linewidth}
        \centering
        \includegraphics[width=\linewidth]{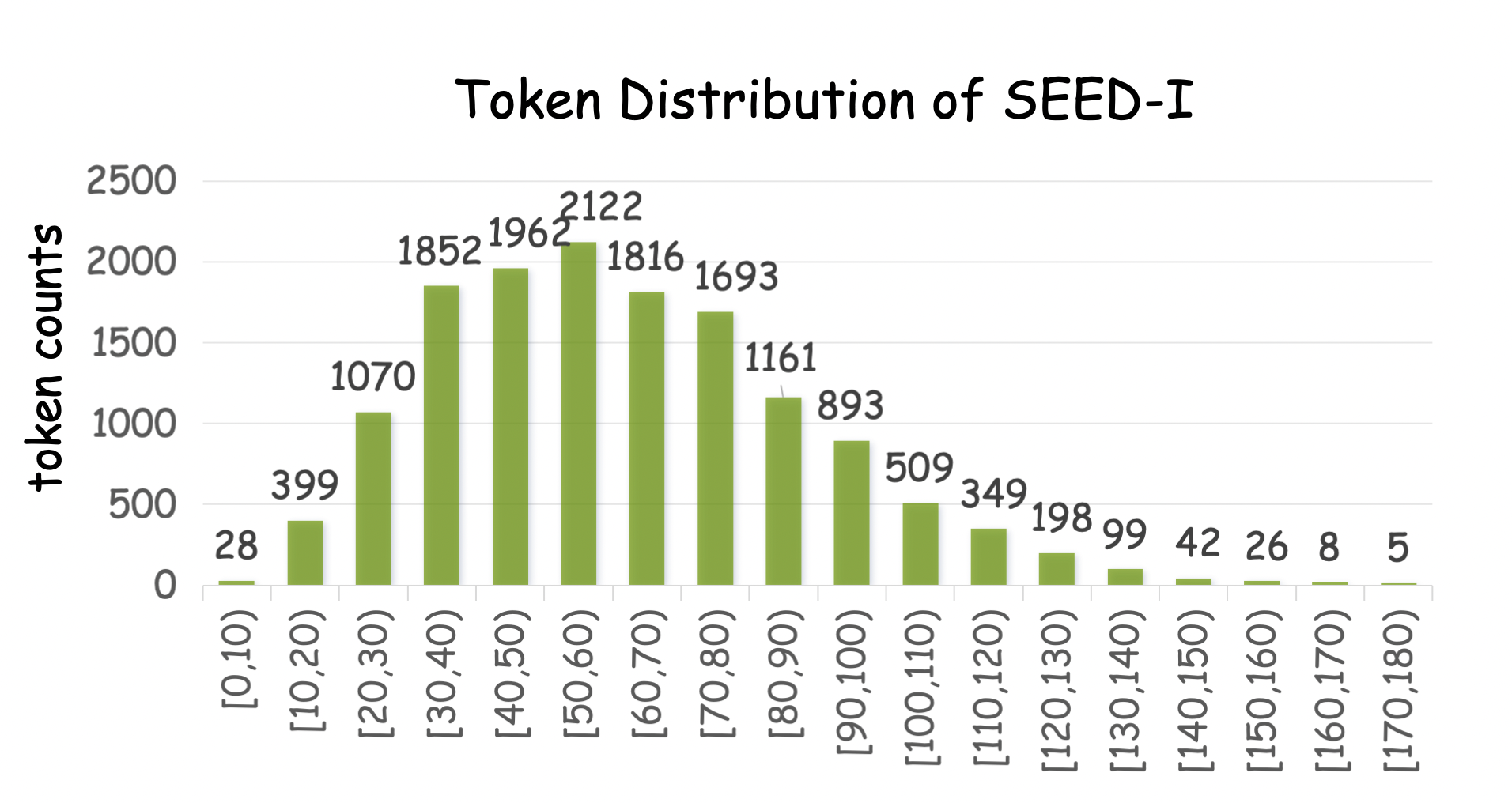}
        \centerline{(e) Token count analysis on the SEED-I benchmark.}
    \end{minipage}
    \vspace{1em}

    \begin{minipage}{\linewidth}
        \centering
        \includegraphics[width=\linewidth]{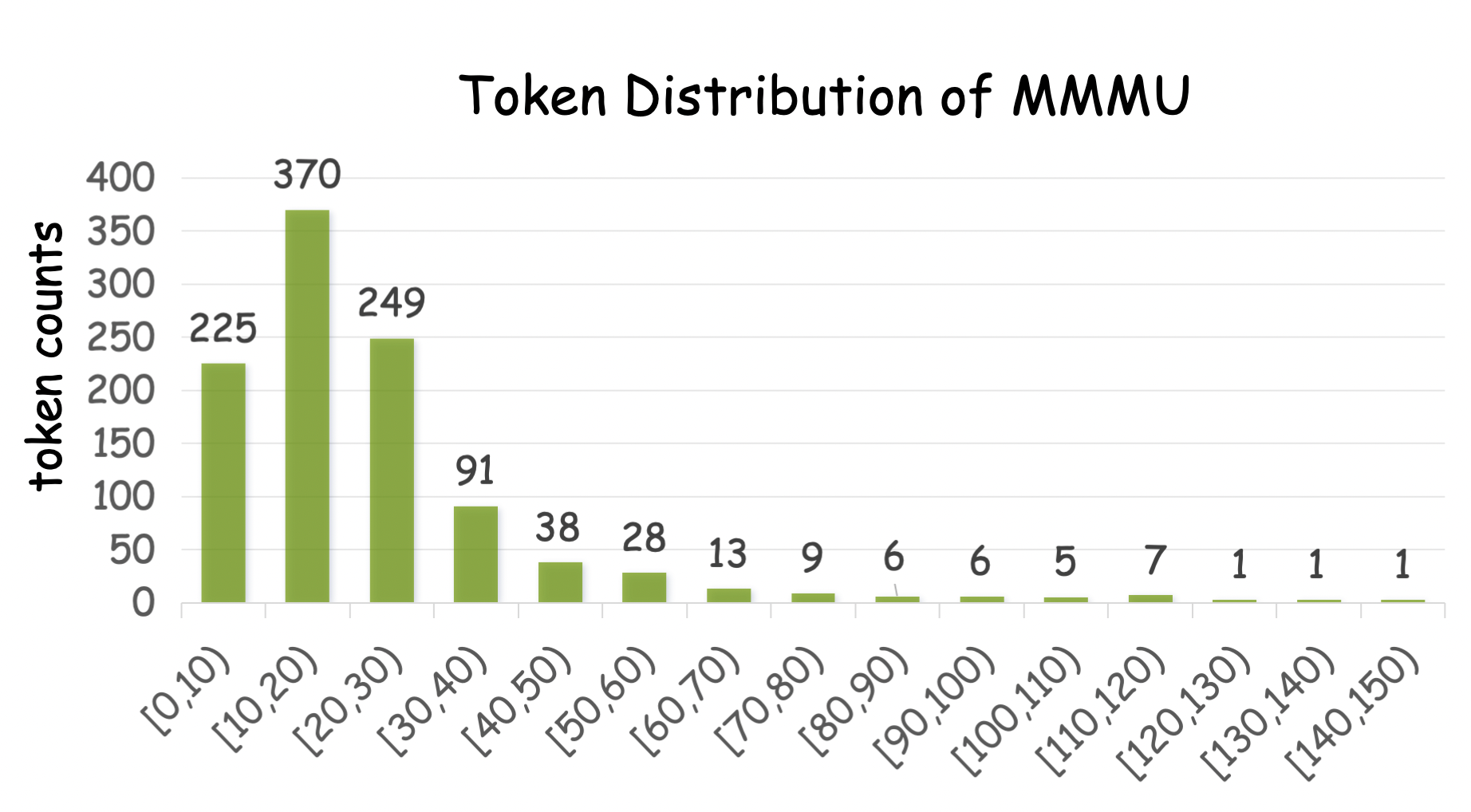}
        \centerline{(f) Token count analysis on the MMMU benchmark.}
    \end{minipage}
    \vspace{1em}

    \captionof{figure}{Detailed token count analysis across the six evaluation benchmarks. Each subfigure shows the results for a specific dataset.}
    \label{fig:full_token_analysis}
\end{center}

\section{More Token Merging Examples}

In this section, we give more token merging examples which include both things and stuff. Each caption briefly describe the corresponding object.

\begin{center}
    \begin{minipage}{0.48\linewidth} 
        \centering
        \includegraphics[width=\linewidth]{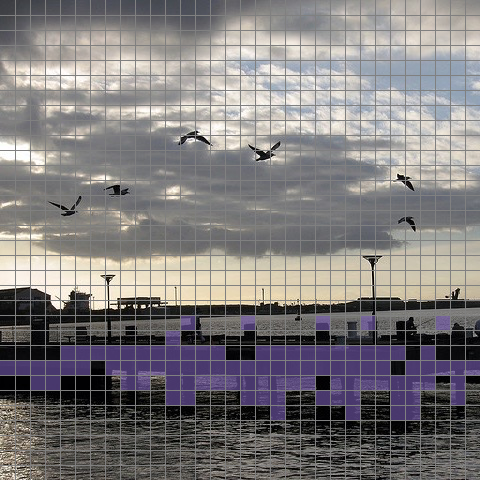}
        \centerline{(a) Bridge}
    \end{minipage}
    \hfill 
    \begin{minipage}{0.48\linewidth}
        \centering
        \includegraphics[width=\linewidth]{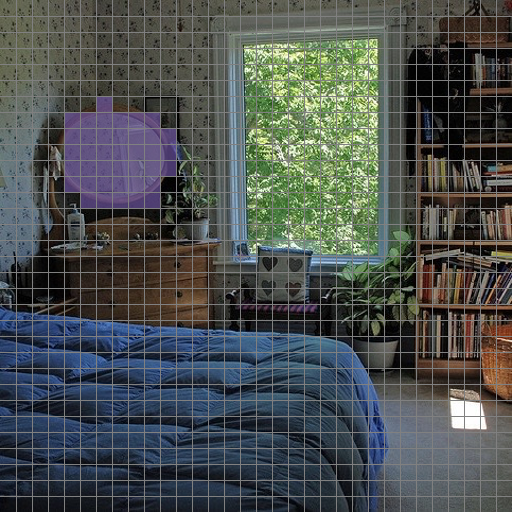}
        \centerline{(b) Mirror}
    \end{minipage}
    \vspace{1.5em} 

    \begin{minipage}{0.48\linewidth}
        \centering
        \includegraphics[width=\linewidth]{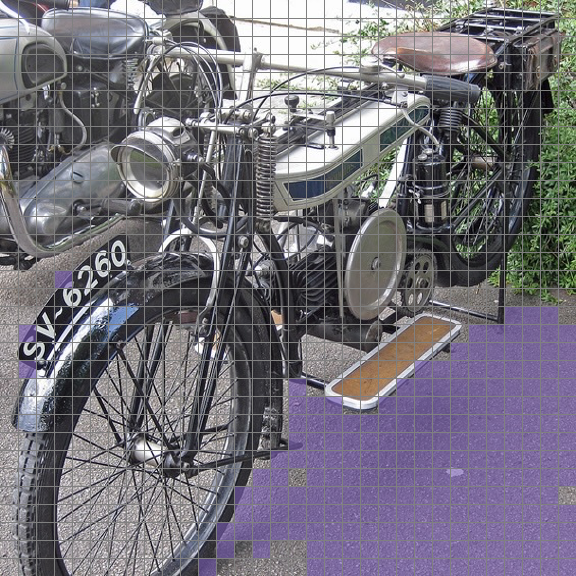}
        \centerline{(c) Ground}
    \end{minipage}
    \hfill
    \begin{minipage}{0.48\linewidth}
        \centering
        \includegraphics[width=\linewidth]{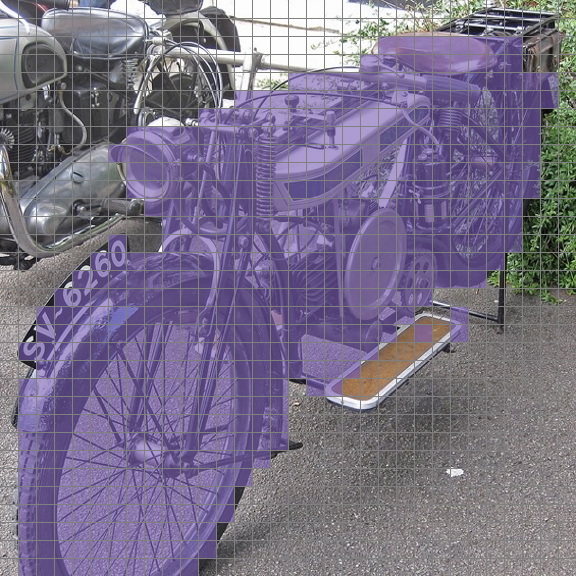}
        \centerline{(d) Bicycle}
    \end{minipage}
    \vspace{1.5em}

    \begin{minipage}{0.48\linewidth}
        \centering
        \includegraphics[width=\linewidth]{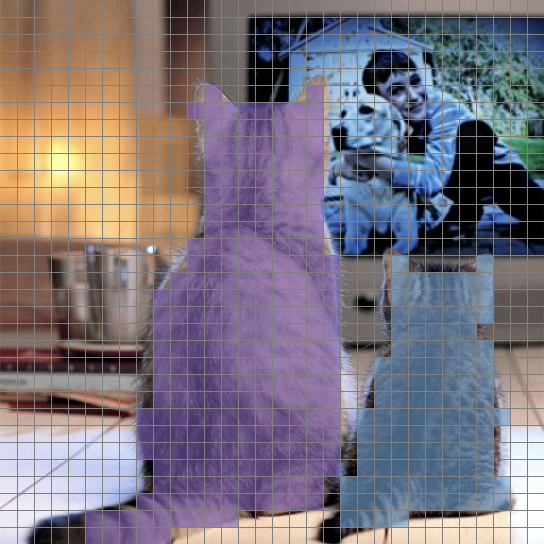}
        \centerline{(e) Two Cats}
    \end{minipage}
    \hfill
    \begin{minipage}{0.48\linewidth}
        \centering
        \includegraphics[width=\linewidth]{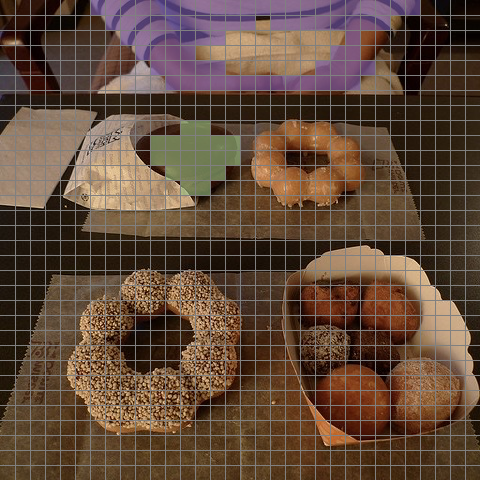}
        \centerline{(f) Hands and Dessert}
    \end{minipage}
    \vspace{1.5em}

    \begin{minipage}{0.48\linewidth}
        \centering
        \includegraphics[width=\linewidth]{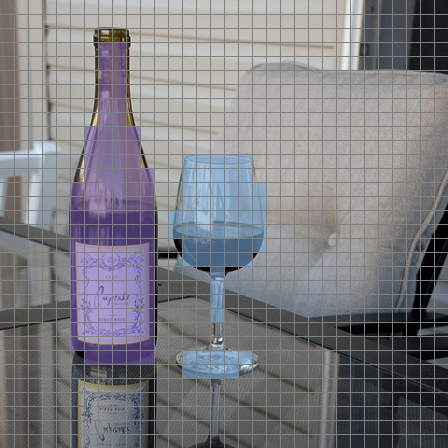}
        \centerline{(g) Bottle and Glass}
    \end{minipage}
    \hfill
    \begin{minipage}{0.48\linewidth}
        \centering
        \includegraphics[width=\linewidth]{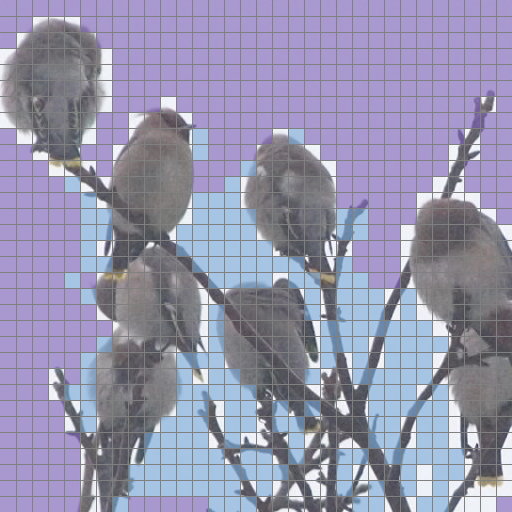}
        \centerline{(h) Snowfield}
    \end{minipage}
    \vspace{1.5em}

    \begin{minipage}{0.48\linewidth}
        \centering
        \includegraphics[width=\linewidth]{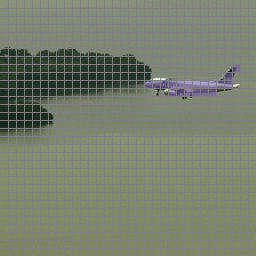}
        \centerline{(i) Plane}
    \end{minipage}
    \hfill
    \begin{minipage}{0.48\linewidth}
        \centering
        \includegraphics[width=\linewidth]{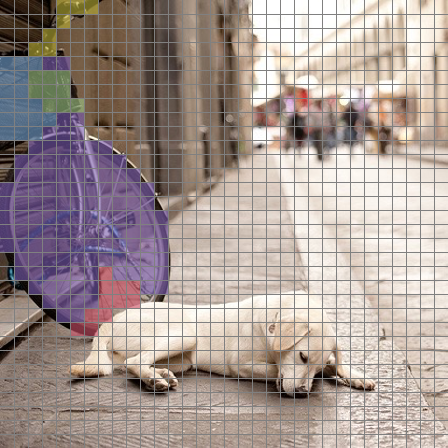}
        \centerline{(j) Bicycle}
    \end{minipage}
    \vspace{1.5em}

    \begin{minipage}{0.48\linewidth}
        \centering
        \includegraphics[width=\linewidth]{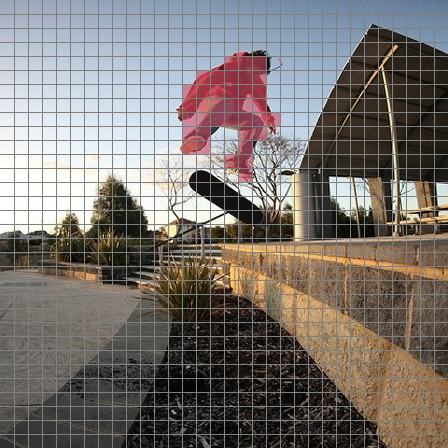}
        \centerline{(k) Skateboarder}
    \end{minipage}
    \hfill
    \begin{minipage}{0.48\linewidth}
        \centering
        \includegraphics[width=\linewidth]{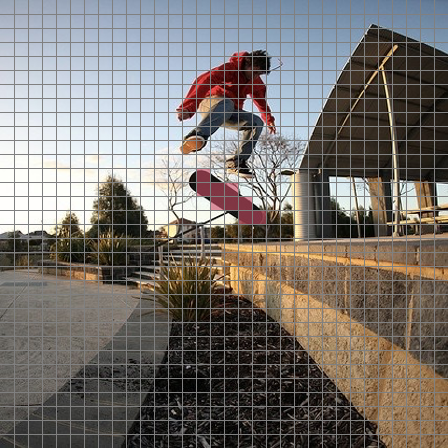}
        \centerline{(l) Skateboard}
    \end{minipage}
    \vspace{1.5em}

    \begin{minipage}{0.48\linewidth}
        \centering
        \includegraphics[width=\linewidth]{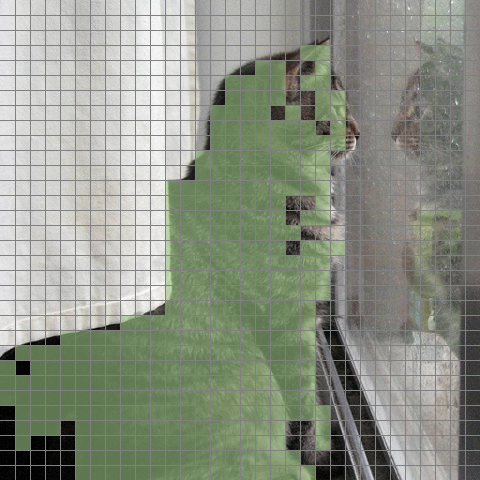}
        \centerline{(m) Cat}
    \end{minipage}
    \hfill
    \begin{minipage}{0.48\linewidth}
        \centering
        \includegraphics[width=\linewidth]{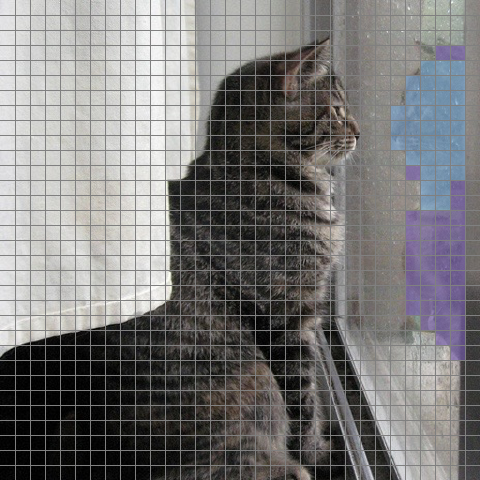}
        \centerline{(n) Cat's Reflection}
    \end{minipage}
    \vspace{1.5em}

    \begin{minipage}{0.48\linewidth}
        \centering
        \includegraphics[width=\linewidth]{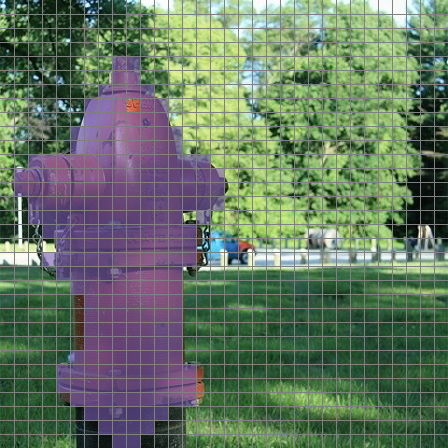}
        \centerline{(o) Hydrant}
    \end{minipage}
    \hfill
    \begin{minipage}{0.48\linewidth}
        \centering
        \includegraphics[width=\linewidth]{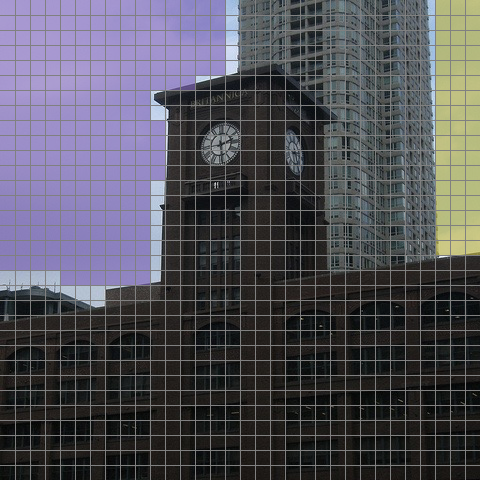}
        \centerline{(p) Sky}
    \end{minipage}
    \vspace{1.5em}

    \captionof{figure}{\textbf{Token Merging Examples.} Tokens with the same color in a image are merged into one. A complex or blocked object may have more than one token after merged. Stuff (e.g., sky, snowfield) is merged equally as things.}
    
\end{center}

\section{Dialogue Comparision}

\begin{figure} [htbp]
    \centering
    \includegraphics[width=\linewidth]{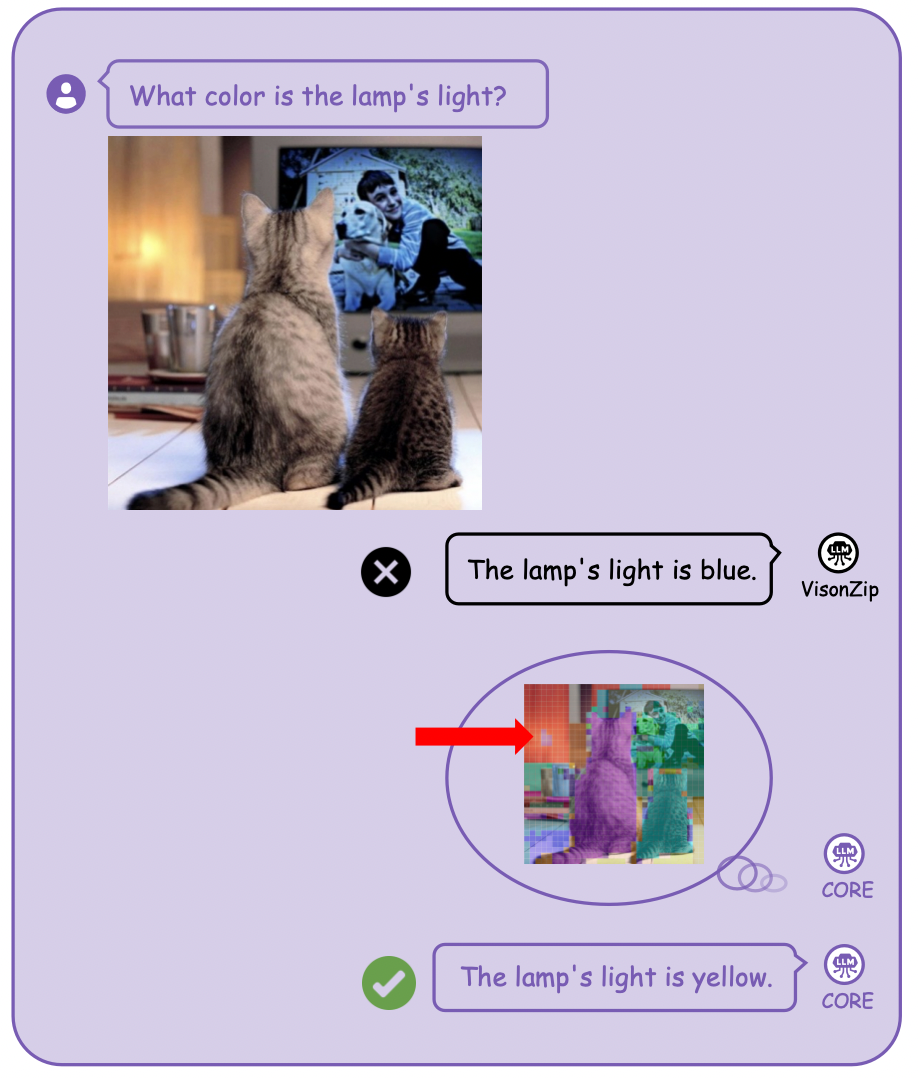}
    \caption{\textbf{Comparison on Background Object Recognition.} The red arrow emphasizes the object-centric token of the lamp, which helps CORE arrive at the correct answer.}
    \label{dialogue_0}
\end{figure}

\begin{figure} [t!]
    \centering
    \includegraphics[width=\linewidth]{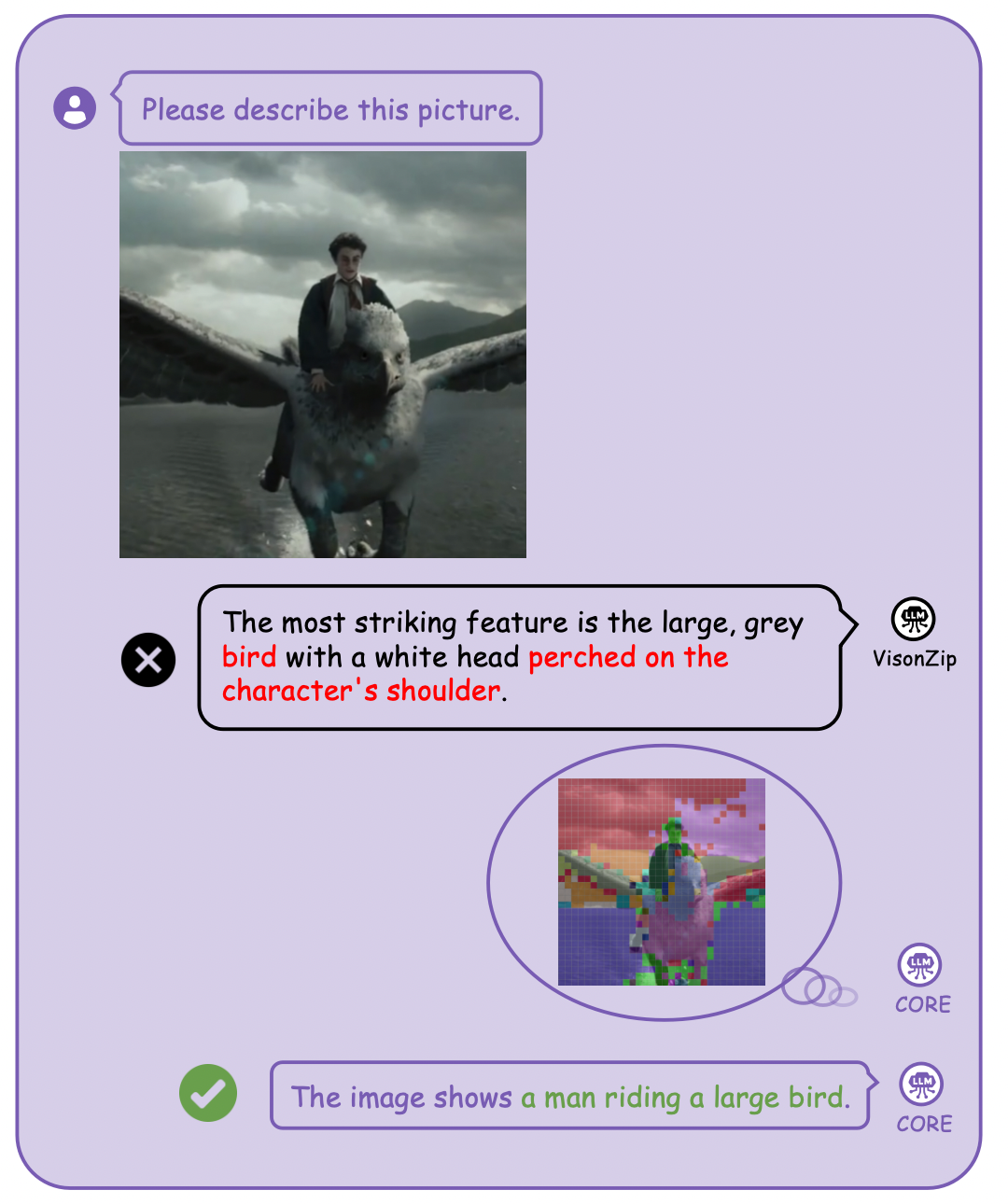}
    \caption{\textbf{Comparison on Objects' Positional Relationship Detection.} Incorrect and correct model responses are highlighted in red and green, respectively.}
    \label{dialogue_1}
\end{figure}

In this section, we compare our CORE model with another token comparison method, VisionZip \citep{yang2024visionzip}, with the same number of retained tokens. We mainly examine the models' ability on background object recognition and objects' positional relationship detection. In Fig.~\ref{dialogue_0} and Fig.~\ref{dialogue_1}, we also show the mask CORE generates internally as CORE's thinking process.

In Fig.~\ref{dialogue_0}, the image is that a large and a small cat are watching TV under a yellow light. CORE produces an object-centric representation for each object in the image, including those in the background, which leads to a correct answer (yellow light). In contrast, other methods may aggressively prune necessary tokens, resulting in an incorrect description (blue light). This comparison demonstrates CORE's comprehensiveness in object recognition, even for background elements.

Fig.~\ref{dialogue_1} gives a comparison of different models' description on objects' positional relationship. The scene is from a movie and it shows the character rides a giant bird over the water. VisionZip gives an incorrect answer that the bird is on the character's shoulder. While based on CORE's centroid-guide sorting strategy, the character's tokens is prior to the bird's. As a result, our model infers that the character rides the bird and gives the correct description. The thinking process shows the complete segmentation mask.

